\documentclass[journal]{IEEEtran}\makeatletter
\def\endthebibliography{%
  \def\@noitemerr{\@latex@warning{Empty `thebibliography' environment}}%
  \endlist
}
\makeatother

\usepackage{graphics}
\usepackage{graphicx}
\usepackage{algorithm}
\usepackage{subfigure}
\usepackage{amsfonts}
\usepackage{amsmath,amssymb}
\usepackage{xcolor}
\usepackage{tikz}
\usetikzlibrary{calc}
\usetikzlibrary{backgrounds,fit}
\usetikzlibrary{matrix,shapes,arrows,positioning,chains}

\usepackage[noend]{algpseudocode}

\begin{document}
\title{A Robust Morphological Approach for Semantic Segmentation of Very High Resolution Images}

\author{Siddharth~Saravanan,~
        Aditya~Challa,~
        and~Sravan~Danda,~\IEEEmembership{Senior Member,~IEEE}
\thanks{Siddharth Saravanan (ssiddharthsaravanan@gmail.com) is with University of California, San Diego, California, 92092, USA. Aditya Challa and Sravan Danda are with Computer Science and Information Systems, BITS-Pilani Goa Campus, Zuari Nagar, 403726, India (aditya.challa.20@gmail.com, sravan8809@gmail.com)}}

%

\maketitle
\begin{abstract}
State-of-the-art methods for semantic segmentation of images involve computationally intensive neural network architectures. Most of these methods are not adaptable to high-resolution image segmentation due to memory and other computational issues. Typical approaches in literature involve design of neural network architectures that can fuse global information from low-resolution images and local information from the high-resolution counterparts. However, architectures designed for processing high resolution images are unnecessarily complex and involve a lot of hyper parameters that can be difficult to tune. Also, most of these architectures require ground truth annotations of the high resolution images to train, which can be hard to obtain. In this article, we develop a robust pipeline based on mathematical morphological (MM) operators that can seamlessly extend any existing semantic segmentation algorithm to high resolution images. Our method does not require the ground truth annotations of the high resolution images. It is based on efficiently utilizing information from the low-resolution counterparts, and gradient information on the high-resolution images. We obtain high quality seeds from the inferred labels on low-resolution images using traditional morphological operators and propagate seed labels using a random walker to refine the semantic labels at the boundaries. We show that the semantic segmentation results obtained by our method beat the existing state-of-the-art algorithms on high-resolution images. We empirically prove the robustness of our approach to the hyper parameters used in our pipeline. Further, we characterize some necessary conditions under which our pipeline is applicable and provide an in-depth analysis of the proposed approach.  
\end{abstract}


%
\IEEEpeerreviewmaketitle

\section{Introduction}
\label{sec:intro}

\begin{figure*}[h]
    \subfigure[]{\label{fig:1a}\includegraphics[scale=0.033]{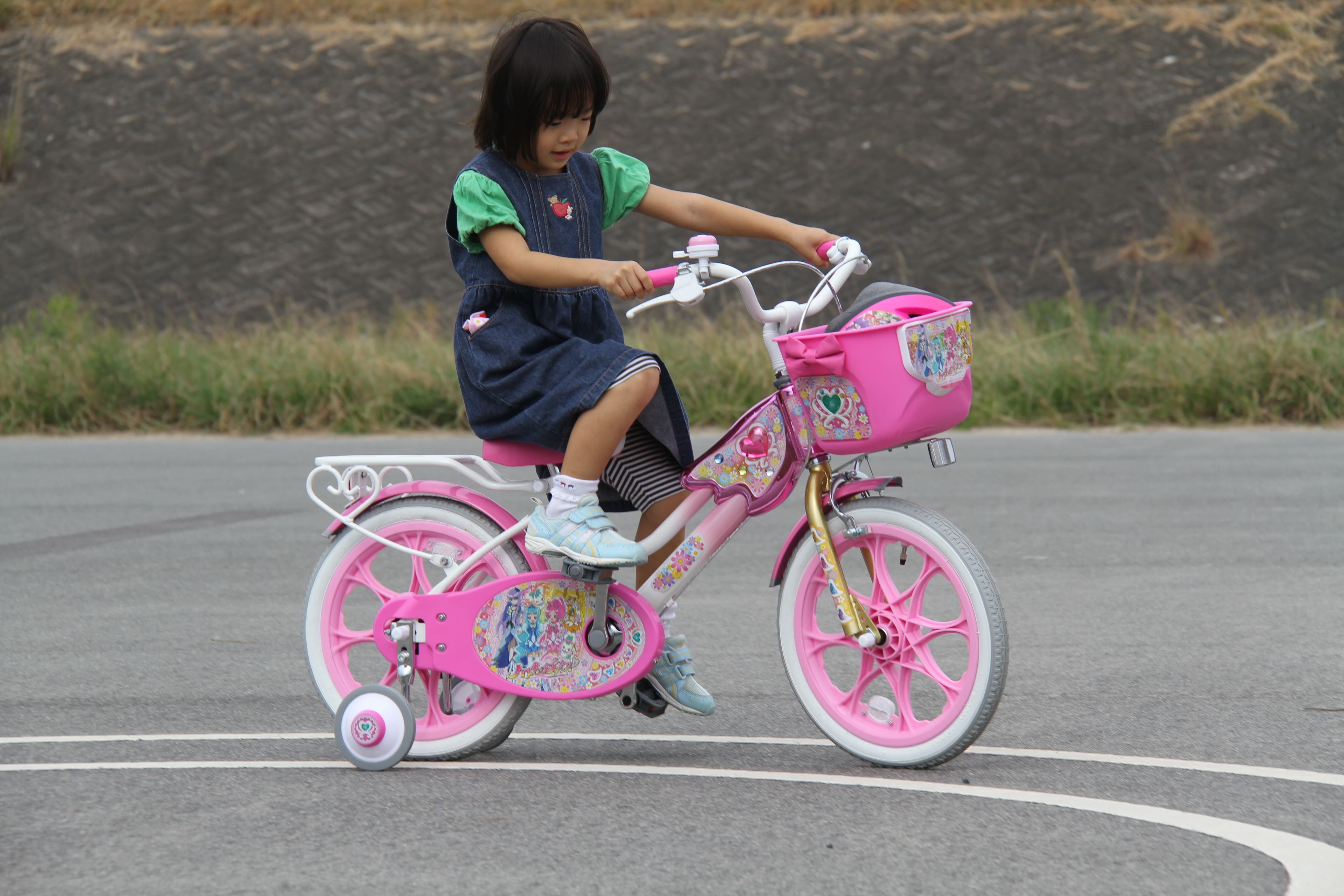}}
    \subfigure[]{\label{fig:1b}\includegraphics[scale=0.033]{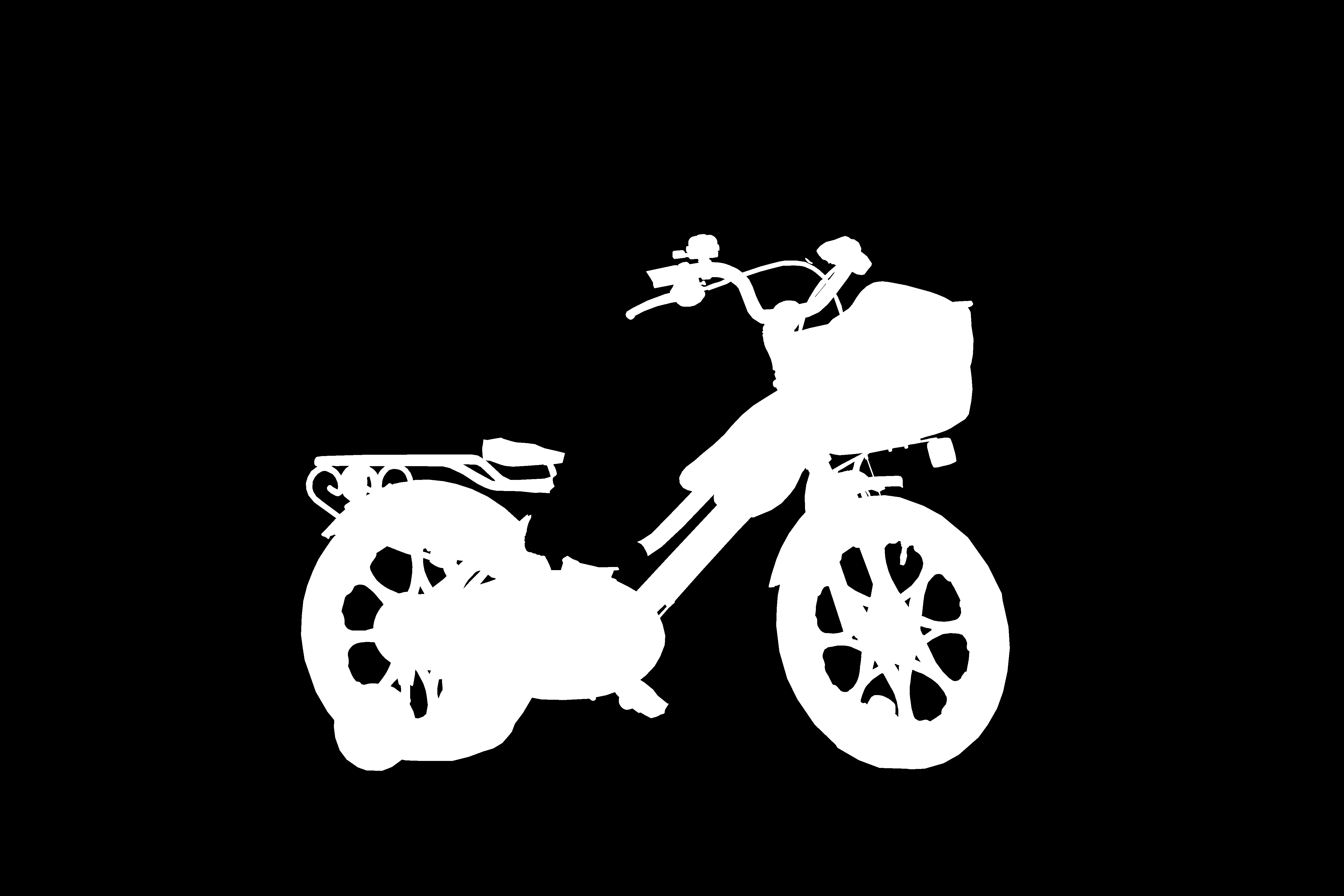}}
    \subfigure[]{\label{fig:1c}\includegraphics[scale=0.033]{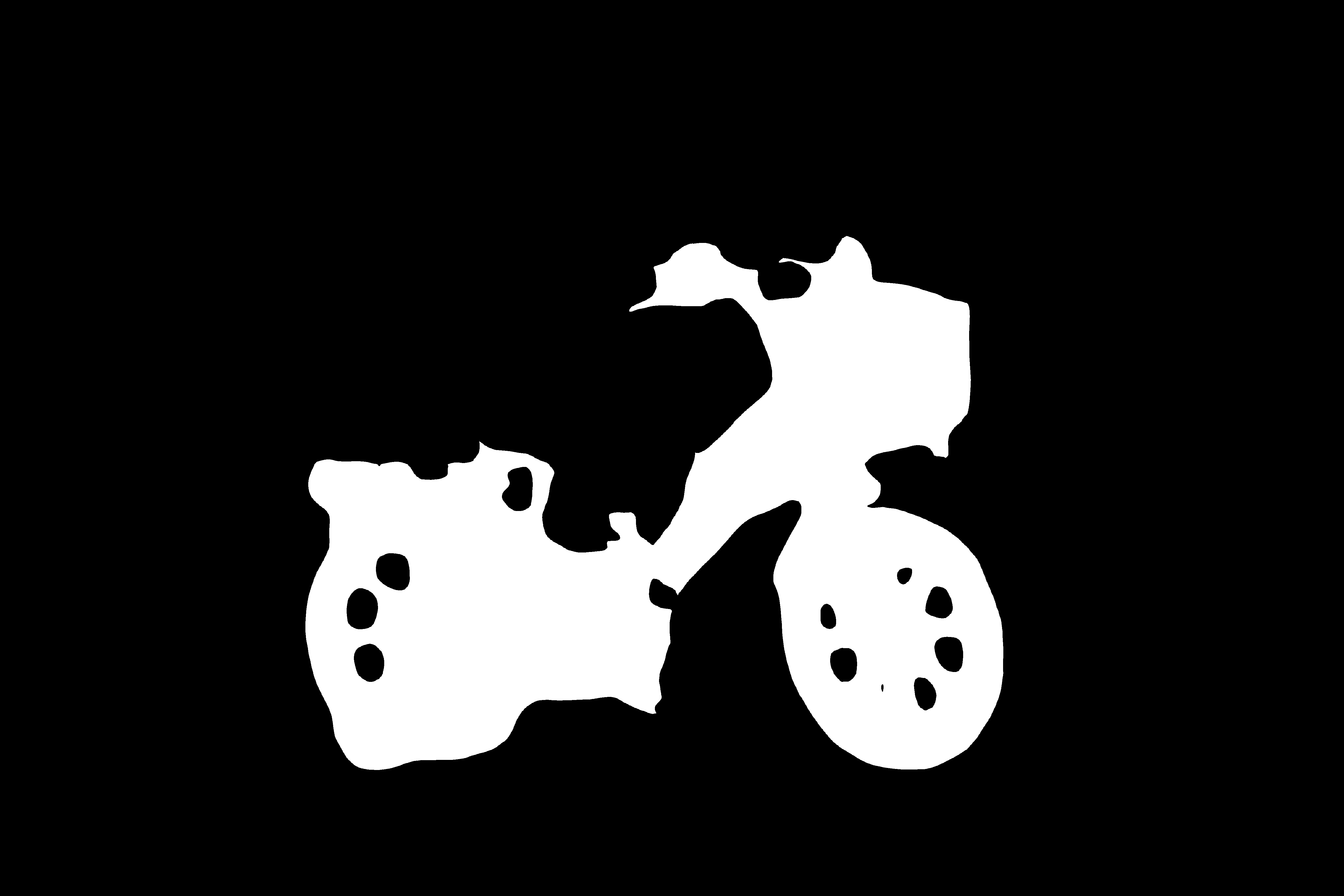}}\\
    \subfigure[]{\label{fig:1d}\includegraphics[scale=0.033]{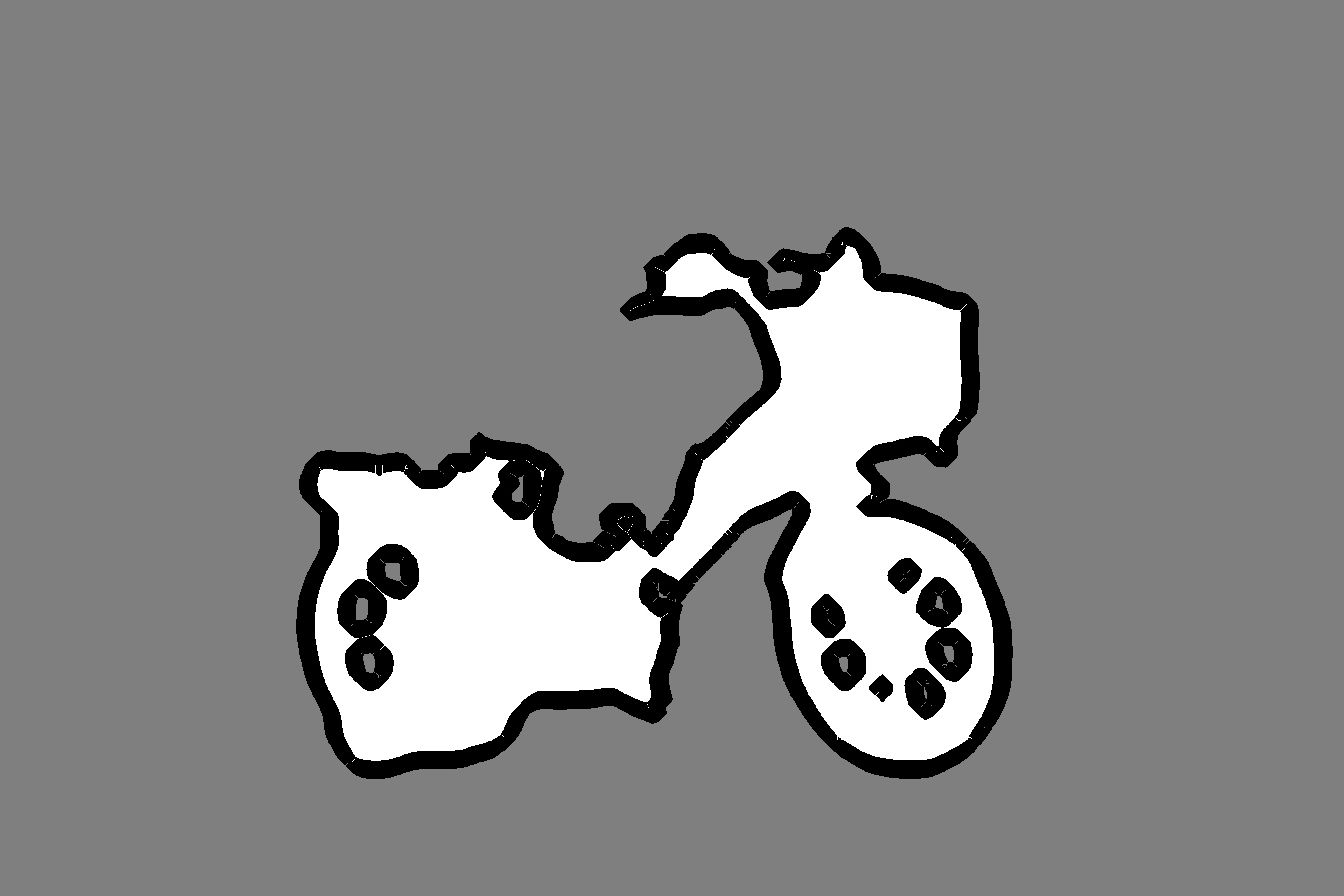}}
    \subfigure[]{\label{fig:1e}\includegraphics[scale=0.033]{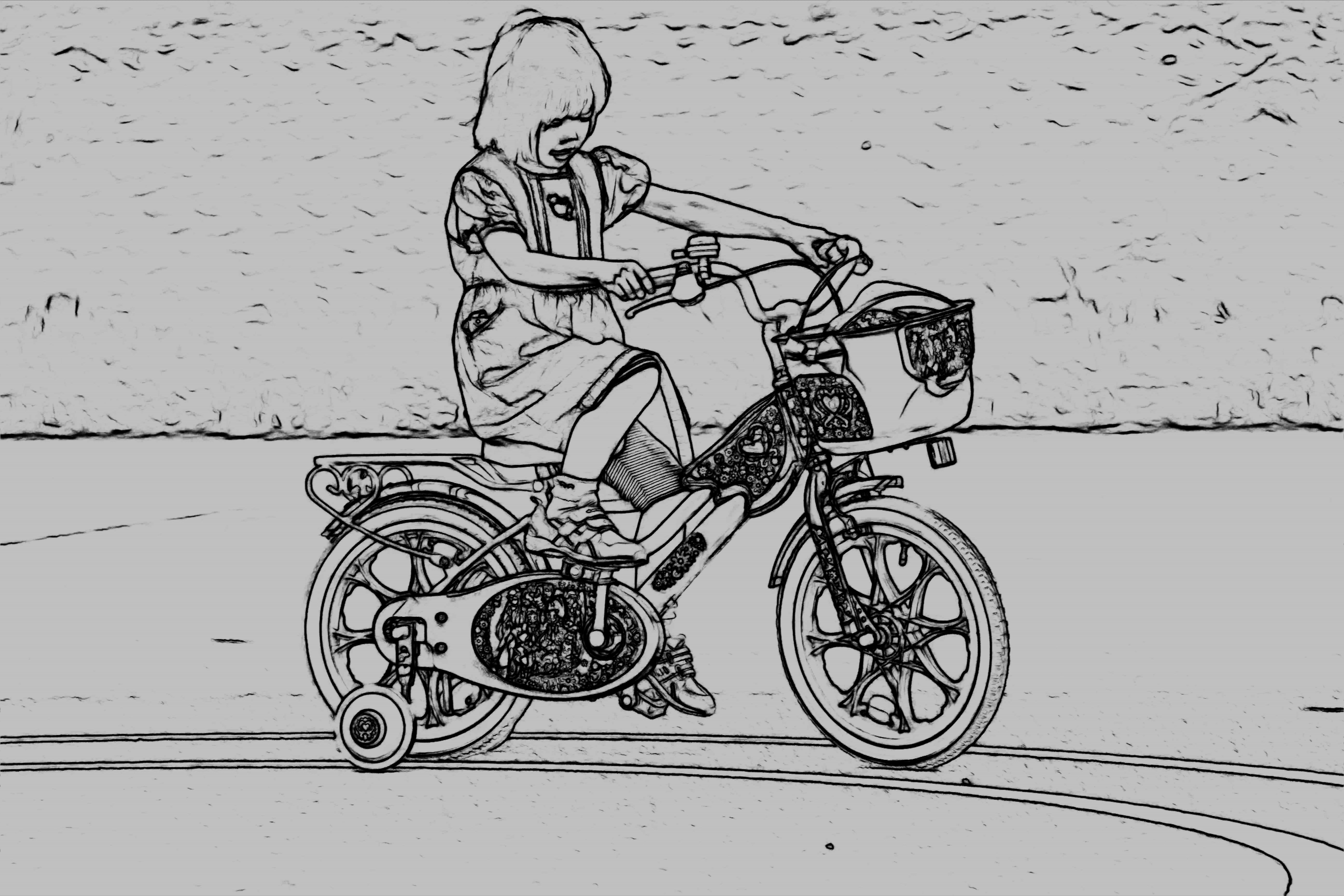}}
    \subfigure[]{\label{fig:1f}\includegraphics[scale=0.033]{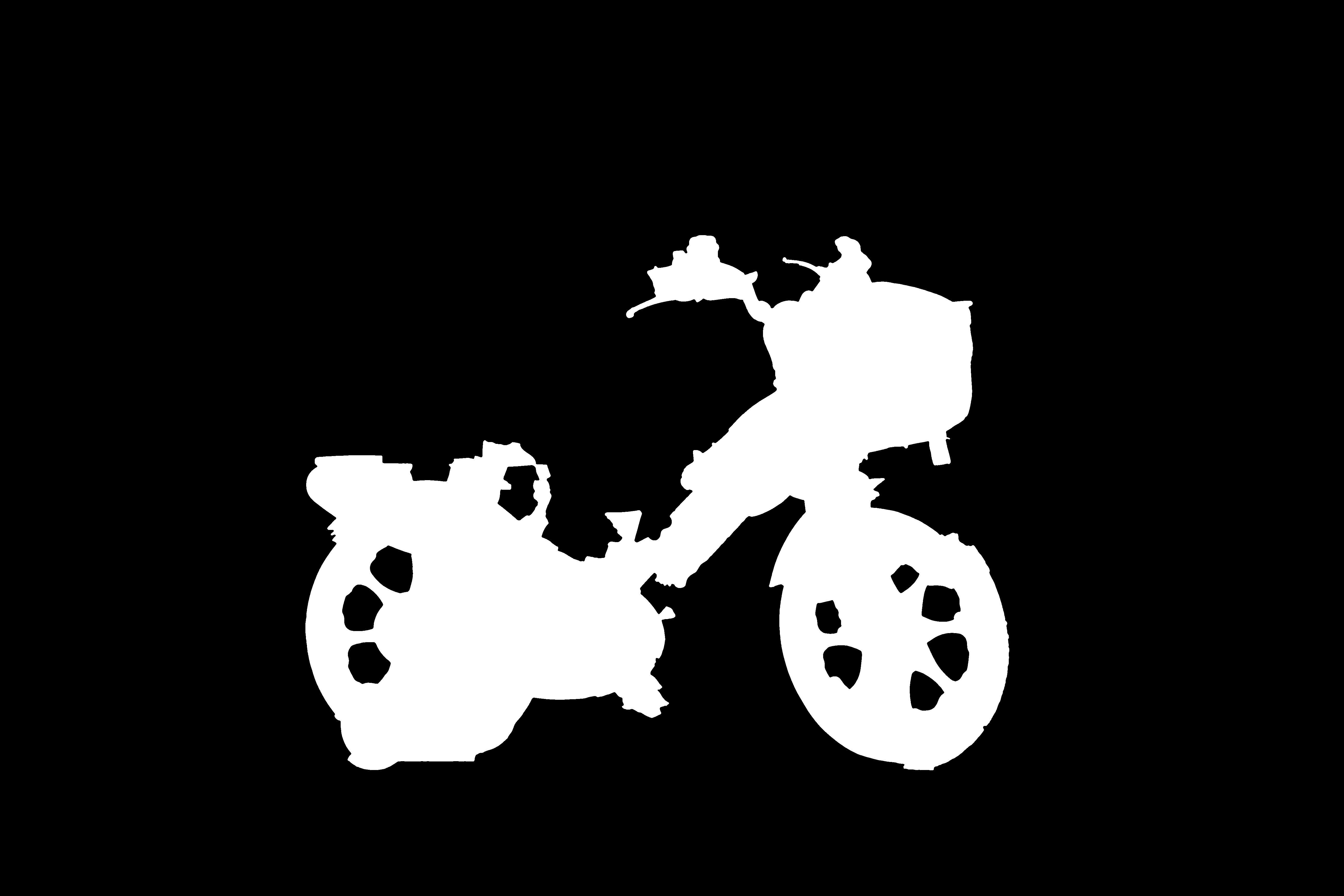}} \\
  \caption{(a) a high resolution image from BIG dataset \cite{cheng2020cascadepsp}. The objective is to identify pixels corresponding to bicycle versus the rest of the pixels. (b) ground truth labels with white representing bicyle and black representing the rest. (c) estimated labels from upsampling the low resolution predictions. (d) The seeds selected by our pipeline by discarding estimated labels closer to object boundaries using a combination of morphological operators. Here white and gray correspond to the pixels that we are certain of, of bicycle and background respectively, and black corresponds to uncertain pixels. (e) edge weights constructed from a learned gradient \cite{dollar2013structured} on the high resolution image. The edge weights are visualized using 2-D cubical complexes \cite{kovalevsky1989finite}. (f) refined labels obtained by random walker \cite{grady2006random} using (d) and (e) as inputs.}
  \label{fig:MorphologicalUHDSemSeg}
\end{figure*}


In the past decade, there has been a steep rise in the development of computer vision algorithms to address tasks such as image segmentation \cite{paszke2016enet,howard2019searching}, semantic image segmentation \cite{minaee2021image,long2015fully,cai2021multi,ding2020semantic,zhao2017pyramid,lin2017refinenet,guo2019degraded}, instance segmentation \cite{minaee2021image}, object detection \cite{zhang2021weakly}, object localization \cite{zhang2021weakly} etc. Computer vision algorithms evolved from those built on designing hand-crafted features to designing deep neural network architectures. The improvement in computational capabilities is one of the major factors that led to this trend. The advancement in high-resolution cameras allows one to utilize richer information making it theoretically possible to obtain an enhanced performance on such tasks. However, architectures designed for low-resolution images do not extend to high-resolution images. This is due to hardware limitations such as usage of GPU memory, scarcity of good-quality training data. In this article, we provide a simple yet robust solution to efficiently extend the semantic segmentation models built for low-resolution images to high-resolution images. 

The aim of image semantic segmentation is to obtain a meaningful semantic label (for example, in a natural image, labels may be `sky', `car', `cat', etc) for every pixel in a given image. For semantic segmentation of low-resolution images, there is an extensive literature available and most of involve deep neural networks. Guidance-based networks \cite{wu2020cgnet,zhang2022guided}, attention-based networks \cite{he2022efficient,li2022attention}, semi-supervised and weakly-supervised methods \cite{yi2021learning,ke2022three,zhou2021group}, gated network \cite{geng2021gated}, coarse-to-fine refinement \cite{jing2019coarse}, random forests on learned representations \cite{kang2019random}, granularity-based \cite{he2021mgseg}, few-shot-based \cite{liu2021harmonic}, flow-based \cite{li2020semantic}, transformer-based \cite{yuan2019segmentation,chen2022vision} etc are some of the existing approaches. These algorithms yield satisfactory results on a variety of datasets such as Microsoft COCO \cite{lin2014microsoft}, PASCAL VOC 2012 \cite{everingham2015pascal}, Cityscapes \cite{cordts2016cityscapes}, ADE20k \cite{zhou2017scene}, Kinect \cite{silberman2012indoor} etc. However, it is impractical to extend these models to high resolution images due to computational constraints. Working with downsampled images and upsampling the results would lead to boundary artifacts and loss of finer structures. Alternatively, one could work with overlapping cropped images but the global information would be lost with such an analysis (see \cite{chen2019collaborative} for details).

Several of the recent approaches \cite{li2022attention,chen2019collaborative,cheng2020cascadepsp,dai2021faster} involve design of neural network architectures that can combine local and global information from downsampled images and multiple cropped patches of high resolution images in various ways. Other recent approaches \cite{jin2020foveation,jin2021learning} involve design of neural networks that are capable of identifying important patches in the high resolution images and are based on the idea of non-uniform sampling of the pixels to reduce the computational burden. Although these methods alleviate the memory issues for model training, most of them require ground truth annotations of the high resolution images. Also, they contain a lot of hyper parameters which are difficult to tune. Further, it is known from \cite{chen2019collaborative} that some of these methods have poor performance when there is a class imbalance  i.e. when the relative proportion of pixels corresponding to different semantic labels within the image are extremely low or high. 

Existing state-of-the-art semantic segmentation methods on high resolution images do not exploit the information provided by semantic segmentation algorithms on low resolution images efficiently. To elaborate, it is intuitively clear that an upsampling of the low resolution image: 1) contains the same set of semantic labels as that of the high resolution image, 2) there is a bijective correspondence between every semantic instance in the ground truth high resolution image and the semantic instances in the upscaled labels, and 3) majority of the incorrect predictions of the upsampling results are closer to object boundaries (see Fig \ref{fig:UpscaleErrorsAndBoundaries}). Thus, given a semantic segmentation algorithm for low resolution images, it is redundant to train a model to learn the semantic labels from scratch. Refining the labels at object boundaries (see Fig \ref{fig:MorphologicalUHDSemSeg} for an illustration) is an efficient way to obtain semantic segmentation.

Our method is built on the key observations mentioned in the previous paragraph. It is similar in spirit to methods that fuse information from high resolution images and the corresponding low resolution images but does not require ground truth annotations of the high resolution images. We model images as grid graphs, and our pipeline broadly consists of three blocks:
\begin{enumerate}
\item obtaining high quality seeds with morphological post-processing of semantic segmentation results on low-resolution images, 
\item obtaining edge weights using simple learned gradients on high-resolution counterparts, and 
\item propagating the seeds using a seeded segmentation method to obtain label estimates. 
\end{enumerate}
 The reader may refer to Fig \ref{fig:Schematic} in Sec \ref{subsec:Core} for a visual representation of our pipeline. 

The key contribution of this article is a pipeline\footnote{https://github.com/SiddharthSaravanan/UHDImageSegmentation/ contains the source code} to extend semantic segmentation algorithms on low-resolution images to high-resolution images that  

\begin{enumerate}
\item does not require ground truth annotations of high-resolution images to train,
\item obtains comparable results to the existing state-of-the-art algorithms,
\item is robust to the inputs given to the algorithm,
\item has minimal hyper-parameters and are easy to tune, 
\end{enumerate}

Further, we briefly describe the relation between our method and the existing works. The rest of the article is organized as follows: In sec \ref{sec:Literature}, existing methods are briefly surveyed and the merits and demerits are discussed. In sec \ref{sec:Main}, based on standard evaluation measures, the semantic segmentation problem is formally described as an optimization problem. This is followed by a detailed description of our pipeline. The relation of our methods to existing methods are then discussed. In sec \ref{sec:Experiments}, some qualitative and quantitative comparisons to the existing state-of-the-art are made to establish that the results obtained by our pipeline are comparable. Next, the robustness of our method to the inputs and hyperparameters are empirically verified. The section is concluded by describing some limitations of our pipeline. In sec \ref{sec:Conclusions}, the results from the article are summarized and some potential directions to pursue are briefly described.

\section{Existing Literature and Related Work}
\label{sec:Literature}
The existing literature can be classified broadly into three overlapping categories namely multi-scale approaches, approaches based on fusion of local and global information, and approaches based on non-uniform sampling. These methods are briefly described below:

\textbf{CascadePSP} \cite{cheng2020cascadepsp} uses two refinement modules, one on cropped high-resolution images and the other on downsampled image as a whole. The core idea of cascadePSP is to refine and correct local boundaries at several scales as much as possible and merge the information at all the scales. CascadePSP does not require ground truth annotations of the high-resolution images for training. Also, cascadePSP is capable of handling segmentation without any restrictions on the relative sizes of the objects. However, the model is complex and is not straightforward to use.

\textbf{Collaborative global-local networks} (GLNet) \cite{chen2019collaborative} is yet another multi-scale approach using a local branch and a global branch. The local branch processes cropped-high resolution images and the global branch processes a downsampled image as a whole. The information obtained from the local patches and global patches are then fused using mutually regularized feature maps. GLNet is also capable of handling class imbalance i.e. irrespective of the relatives sizes of the objects, it works well. However, the architecture involves redundant parameters.

\textbf{Faster patch proposal network} (Faster-PPN) \cite{dai2021faster}, a faster version of patch proposal network \cite{wu2020patch} is yet another algorithm based on global-local framework using local and global branches. The key idea of this approach is to optimize for the number of local patches that needs to be processed. Such an optimization reduces the computational complexity.

\textbf{Foveation} \cite{jin2020foveation} is a recent approach inspired by the way humans annotate large images. This algorithm generates importance weights to identify the most informative patches. The segmentation module then takes as input, patches with varied resolutions depending on the importance. This algorithm achieves high-quality results on a popular remote sensing segmentation dataset \cite{demir2018deepglobe}.

\textbf{Deformed downsampling} \cite{jin2021learning} is inspired from the way humans visualize images. The idea is to identify high resolution details in patches that contain important boundary details that contribute to enhanced image segmentation and ignore the high resolution details of other patches. Essentially, this boils down to learning an optimal non-uniform downsampling of an image subject to memory constraints. This approach is flexible enough to be integrated into existing semantic segmentation architectures.

\textbf{Locality aware contextual correlation} \cite{li2021contexts} processes local patches and learns the contextual features between features of neighbouring local patches with a correlation module. The feature information is then fused to combine local-context information to improve local segmentation.

\textbf{Meticulous Net} \cite{yang2020meticulous} introduces a novel encoder-decoder network to refine segmentation at object boundaries. This architecture uses PSPNet. ResNet50, and MobileNetV3 as backbones.

\textbf{Progressive semantic segmentation} \cite{huynh2021progressive} is a multi-scale framework that progressively refines image segmentations at increasing resolutions.

\textbf{Deepstrip} \cite{zhou2020deepstrip} is based on refining the predicted boundaries. The key idea is to obtain potential boundary pixels. The features of only these pixels are learnt using a regularization-based loss and the final boundary pixels are identified from the potential candidates.


Among the recent state-of-the-art algorithms, apart from CascadePSP \cite{cheng2020cascadepsp}, all the architectures require ground truth annotations of the high resolution images. This is a major drawback as it is very hard to obtain ground truth annotations.

\section{From Low-Resolution to High-Resolution Semantic Image Segmentation}
\label{sec:Main}

\begin{figure*}[h]
    \subfigure[]{\label{fig:2a}\includegraphics[scale=0.292]{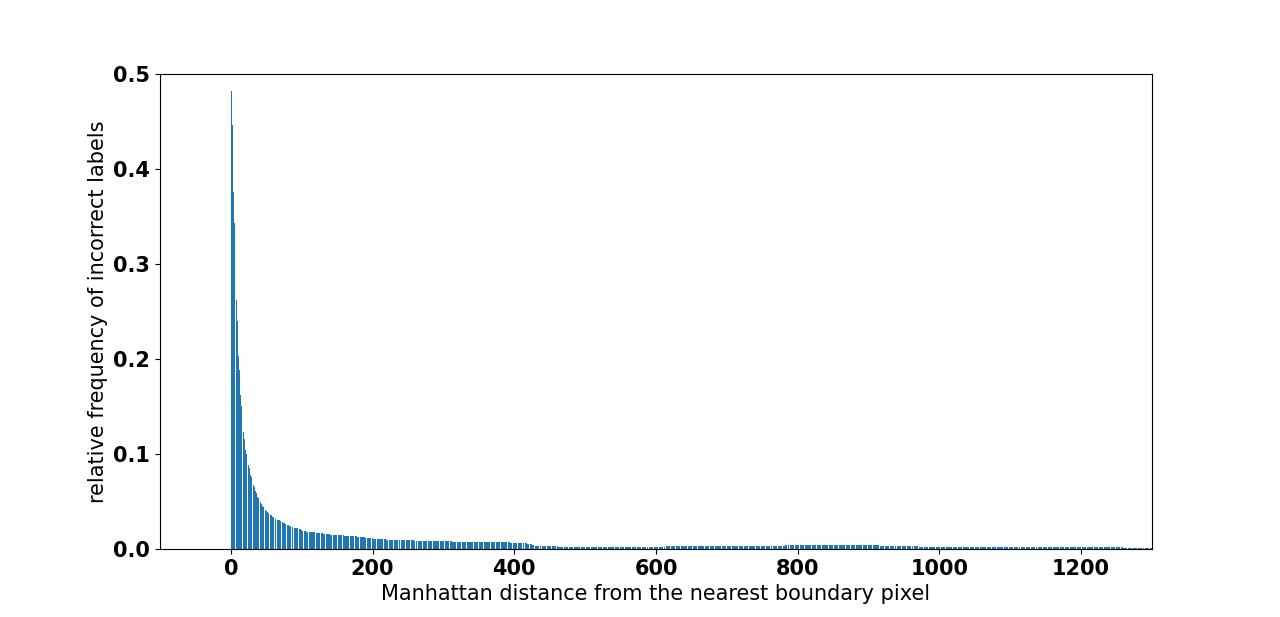}}
    \subfigure[]{\label{fig:2b}\includegraphics[scale=0.292]{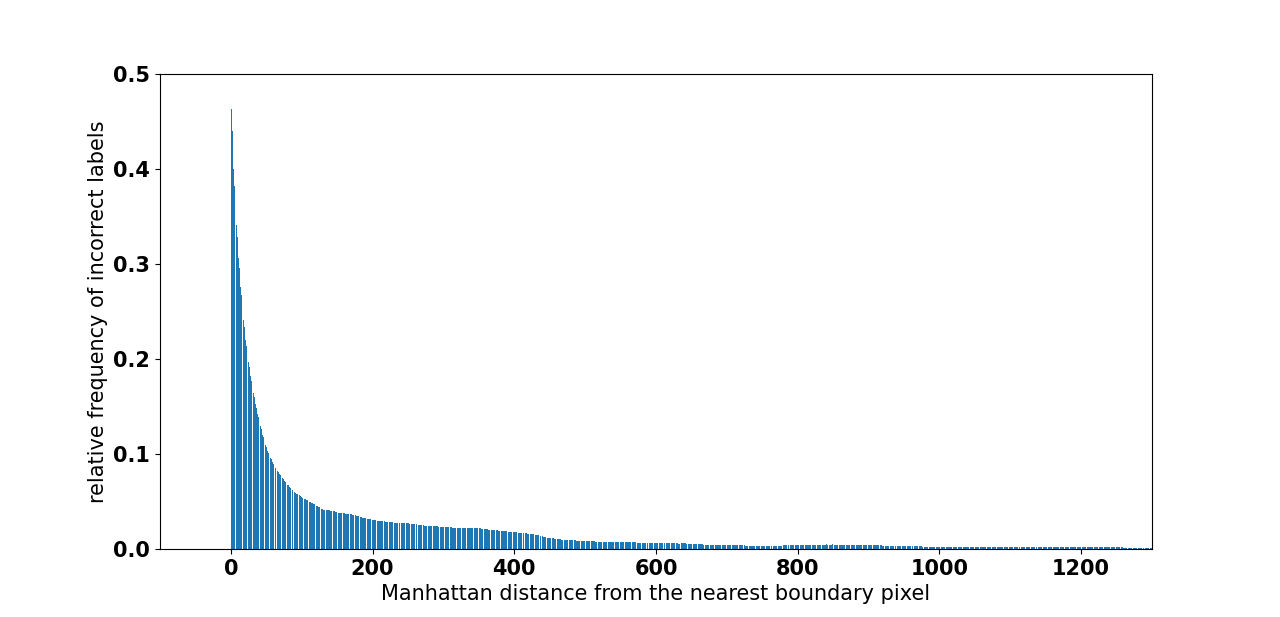}} \\
  \caption{Histograms containing the relative frequencies of incorrect labels as a function of Manhattan distance from object boundaries. (a) corresponds to those obtained by upsampling the estimates of DeepLabV3+ \cite{chen2018encoder}, and (b) corresponds to those obtained by upsampling the estimates of FCN-$8$ \cite{long2015fully}.}
  \label{fig:UpscaleErrorsAndBoundaries}
\end{figure*}

In this section, we formally describe the problem of semantic image segmentation. We then empirically show majority of the incorrect predictions on the upsampled image are near the object boundaries. We then discuss the proposed pipeline and conclude the section by briefly discussing the relation between our method and some existing approaches.

\subsection{Semantic Image Segmentation as an Optimization Problem}
 
The problem of semantic image segmentation can be stated as - given an image, assign a meaningful semantic label to each pixel in the image. Although the problem of semantic image segmentation can be defined for several types of images, we restrict the discussion to $2$D digital images in this article. 

In order to describe semantic image segmentation as an optimization problem, we need some notions. Recall that a $2$D digital image consists of a finite number of pixels arranged as a matrix of size $M \times N$ with $M, N \in \mathbb{Z}^{+}$. Each pixel corresponds to a small physical area. The entries of the matrix are possibly vectors of finite dimension $d$ and the coordinates of each entry represent the average reflectance of light of a particular wavelength. Since we store only discrete values,  a $2$D image $I$ of dimensions can be viewed as a discrete function given by Eq \ref{eq:img}

\begin{equation}
\label{eq:img}
I : \mathbb{Z}_{M,N} \rightarrow \mathbb{Z}^{d}, 
\end{equation}

where $\mathbb{Z}_{M,N} = \{0, 1, \cdots, M-1\} \times \{0, 1, \cdots, N-1\}$. Let $\mathcal{L}$ be a finite set that denotes the set of all possible semantic labels. The ground truth annotations of an image $I$ can thus be represented as another discrete function given by Eq \ref{eq:gtSemanticSeg}

\begin{equation}
\label{eq:gtSemanticSeg}
I_{GT} : \mathbb{Z}_{M,N} \rightarrow \mathcal{L}
\end{equation}

Let $\mathcal{I}$ denote a finite set of images and let $\mathcal{I}_{GT}$ denote the set of corresponding ground truth annotations. For each $I \in \mathcal{I}$, let $D(I)$ denote $\mathbb{Z}_{M,N}$ whenever $I$ is of dimensions $M \times N$. For each $I \in \mathcal{I}$, let $\hat{I}_{GT}$ denote an estimated ground truth annotation and let $\hat{\mathcal{I}}_{GT} = \{\hat{I}_{GT}: I \in \mathcal{I}\}$ denote the set of estimated ground truth annotations of the set of images $\mathcal{I}$. 

For each label $l \in \mathcal{L}$, define true positives $TP_l$ as

\begin{equation}
TP_l = |\{ p \in D(I): \hat{I}_{GT}(p) = I_{GT}(p) = l \ for  \ some \ I \in \mathcal{I}\}|,
\end{equation} 

false positives $FP_l$ as
\begin{equation}
FP_l = |\{ p \in D(I): \hat{I}_{GT}(p) = l \neq  I_{GT}(p) \ for  \ some \ I \in \mathcal{I}\}|,
\end{equation} 

false negatives $FN_l$ as
\begin{equation}
FN_l = |\{ p \in D(I): \hat{I}_{GT}(p) \neq l = I_{GT}(p) \ for  \ some \ I \in \mathcal{I}\}|,
\end{equation} 

and the measure intersection-over-union as $IoU_l$ as
\begin{equation}
IoU_l = \frac{TP_l}{TP_l + FP_l + FN_l}
\end{equation}

Then the problem of semantic image segmentation can be stated as: given a finite set of images $\mathcal{I}$, and corresponding ground truth annotations $\mathcal{I}_{GT}$, find a set of estimates $\hat{\mathcal{I}}_{GT}$ of the ground truth annotations such that the overall intersection over union score function, $IoU_{overall}$  defined by Eq \ref{eq:AvgIoU} is maximized.

\begin{equation}
\label{eq:AvgIoU}
IoU_{overall} = \frac{\sum_{l \in \mathcal{L}}TP_l}{\sum_{l \in \mathcal{L}}\left(TP_l + FP_l + FN_l\right)}
\end{equation}

The ground truth annotations of images i.e. $\mathcal{I}_{GT}$ might be partially available or completely unavailable while training the model. 

\subsection{Useful information from low resolution semantic segmentation}

Recall from Sec \ref{sec:intro} that the basis for our approach is the hypothesis that the incorrect predictions of the upsampled results are near the object boundaries. Here we perform an experiment to validate this hypothesis. We downsample $50$ images and the corresponding ground truth annotations of the BIG dataset \cite{cheng2020cascadepsp} such that the long axis has $512$ pixels while maintaining the aspect ratio. We use two standard semantic segmentation models namely DeepLabV3+ \cite{chen2018encoder}, and FCN-$8$ \cite{long2015fully} to obtain estimates of the semantic labels on the downscaled images. Using each of these models, estimates are upscaled using nearest neighbour upsampling to match the resolution of the original images. The relative frequency of incorrectly predicted labels are plotted as histograms as a function of Manhattan distance from the nearest boundary pixel. Note that a pixel is a boundary pixel if at least one its $4$-adjacent neighbours has a different ground truth annotation.

Fig \ref{fig:UpscaleErrorsAndBoundaries} contains the plots of the histograms. It can be clearly seen that both the histograms are heavily skewed towards lower distances. This implies that majority of the incorrect predictions on the upscaled results of a standard semantic segmentation algorithm are near the object boundaries. Conversely, the predicted labels from upscaled estimates that are at a large distance from a boundary pixel serve as good quality seeds. In order to trim the pixels that are closer to object boundaries, one requires a good quality edge detector on the high-resolution images to estimate the boundaries. Thus, a standard semantic segmentation algorithm on the low resolution image, a good quality edge detector on high-resolution image, and a standard seed propagation algorithm are enough to refine the predictions. In this article, we use 1) DeepLabV3+, and FCN-$8$ as algorithms to obtain crude estimates of semantic segmentation on low-resolution images, 2) Dollar gradient \cite{dollar2013structured}, a simple learned gradient algorithm for natural images, and 3) random walker \cite{grady2006random} as the seed propagation algorithm to demonstrate how our pipeline works. 

\subsection{Morphological Approach to High Resolution Semantic Image Segmentation}
\label{subsec:Core}


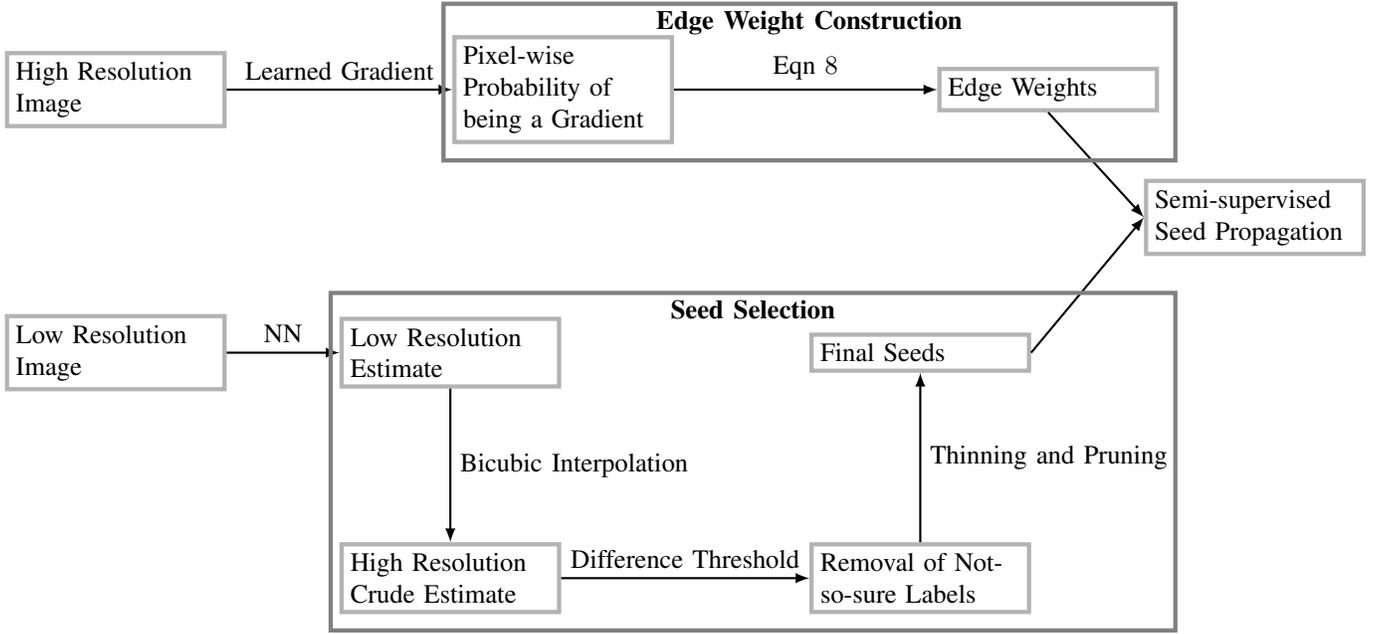
\begin{figure*}
\tikzset{my rectangle/.style={
        draw=black!30, 
        ultra thick, 
        rectangle, 
        text width=0.3*\columnwidth,
        anchor=west, 
        xshift=1.5cm,
        yshift=0.0cm},
}
\tikzset{my arrow/.style={-latex, thick}}

\begin{tikzpicture} 
    \node [my rectangle]                (LRI)       at (0, 0)                  {Low Resolution Image}; 
    \node [my rectangle]                (HRI)       at (0, 3.5)                  {High Resolution Image}; 
    \node [my rectangle, xshift = 1.5cm]                (GradProb)       at (HRI.east)   {Pixel-wise Probability of being a Gradient};    
    \node [my rectangle, xshift = 2.0cm]                (EdgeWt)       at (GradProb.east)   {Edge Weights};     
    \node [my rectangle, yshift= 0.0cm]                (LRE)        at (LRI.east)       {Low Resolution Estimate};
    \node [my rectangle, yshift=-3.0cm]                (HRCE) at (LRI.east)        {High Resolution Crude Estimate};
    \node [my rectangle, xshift = 1.8cm]                (NotSure) at (HRCE.east)        {Removal of Not-so-sure Labels};  
    \node [my rectangle, xshift = 1.8cm]                (Seeds) at (LRE.east)        {Final Seeds};                 
    \node [my rectangle, yshift= 1.8cm]                (RW)       at (Seeds.east)       {Semi-supervised Seed Propagation};

    \draw [my arrow] (LRI.east)      -- node [above] {NN}(LRE.west);
    \draw [my arrow] (LRE.south)         -- node [right] {Bicubic Interpolation}(HRCE.north);
    \draw [my arrow] (HRCE.east)        -- node [above] {Difference Threshold}(NotSure.west);
    \draw [my arrow] (NotSure.north) -- node [right] {Thinning and Pruning}(Seeds.south);
    \draw [my arrow] (Seeds.east)      -- (RW.west);    
    \draw [my arrow] (HRI.east)      -- node [above] {Learned Gradient}(GradProb.west);    
    \draw [my arrow] (GradProb.east)      -- node [above] {Eqn $\ref{eq:EW}$}(EdgeWt.west);
    \draw [my arrow] (EdgeWt.south)         --  (RW.west);

    \node [draw=black!50, ultra thick, 
            fit={(GradProb)
                ([yshift=0.3cm]GradProb.north) 
                ([yshift=-0.1cm]GradProb.south) 
                ([xshift=0.3cm]GradProb.west)
                ([xshift=0.1cm]EdgeWt.east)}
        ] (Edge Weight Construction) {};
    \node [draw=black!50, ultra thick, 
            fit={($(LRE)-(1.0cm,0)$) 
                ([xshift=0.0cm]LRE.west)           
                ([yshift=0.2cm]LRE.north)
                ([yshift=-0.1cm]HRCE.south)  
                ([xshift=1.8cm]Seeds.east)}
        ] (Seed Selection) {};

    \node [anchor=north, font=\bfseries] at (Edge Weight Construction.north) {Edge Weight Construction};
    \node [anchor=north, font=\bfseries] at (Seed Selection.north) {Seed Selection};    

\end{tikzpicture}
\caption{For each pixel, the probability of being a boundary pixel is estimated by applying a learned gradient on the high resolution image. Treating image as a $4$-adjacency graph, for each pair of pixels that share an edge, edge weights are constructed by using a decreasing function of the sum of these probabilities. These edge weights are used as an input for a semi-supervised seed propagation method such as a random walker. On the corresponding low-resolution image, a neural network is applied to obtain crude predictions of labels. The estimated probabilities are then upsampled using bicubic interpolation. From the interpolated probabilities, pixels for which the confidence of the top predicted label among the top two predicted labels is low, are trimmed. Further, thinning is applied iteratively for a preset number of iterations. This is followed by iterative pruning for another preset number of iterations. The thinning and pruning ensures that the pixels close to boundaries for which we are unsure of the estimated labels are trimmed. The remaining pixels at this step are used as seeds for the semi-supervised seed propagation method.}
\label{fig:Schematic}
\end{figure*} 

We model images as $4$-adjacency edge-weighted graphs where nodes represent pixels and edge weights represent similarity between adjacent pixels. Our pipeline consists of three main blocks: 1) seed selection by morphological post processing on the semantic label estimates obtained on low resolution images, 2) construction of edge weights using learned gradients on high resolution images for propagating seeds, and 3) the seed propagation method. A schematic diagram summarizing the pipeline of our method is provided in Fig \ref{fig:Schematic}.

\paragraph{\textbf{Seed selection}} For constructing seeds, we first apply a commonly used algorithm (such as DeeplabV3+, FCN-8 etc) designed to obtain semantic segmentation on low resolution images. We then upscale the probability estimates of each of the semantic classes using a bi-cubic interpolation to obtain crude estimates of the probabilities of all semantic labels at each pixel. From the estimates obtained, we first discard pixels that have absolute difference between probabilities of the top two probable semantic labels below a certain threshold. The exact value of the threshold is a hyperparameter. In the remaining set of pixels, each pixel is viewed as a potential seed for exactly one semantic label. The label of each such pixel is given by the label corresponding to the highest probability in the interpolated probability vector. 

\begin{figure*}[h]
    \subfigure[]{\label{fig:3a}\includegraphics[scale=0.5]{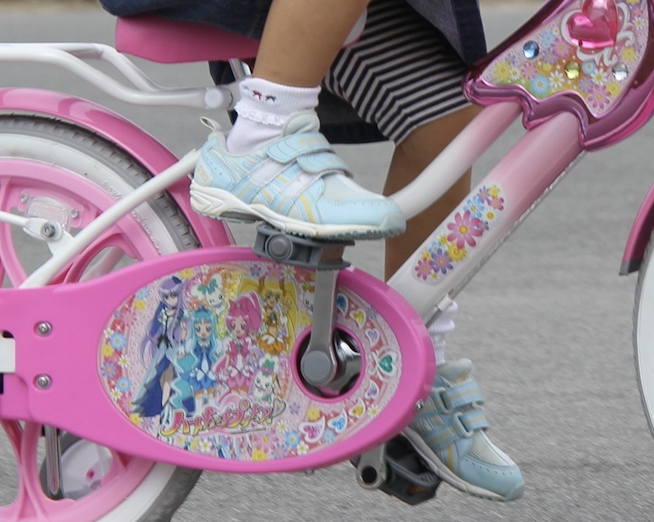}}
    \subfigure[]{\label{fig:3b}\includegraphics[scale=0.5]{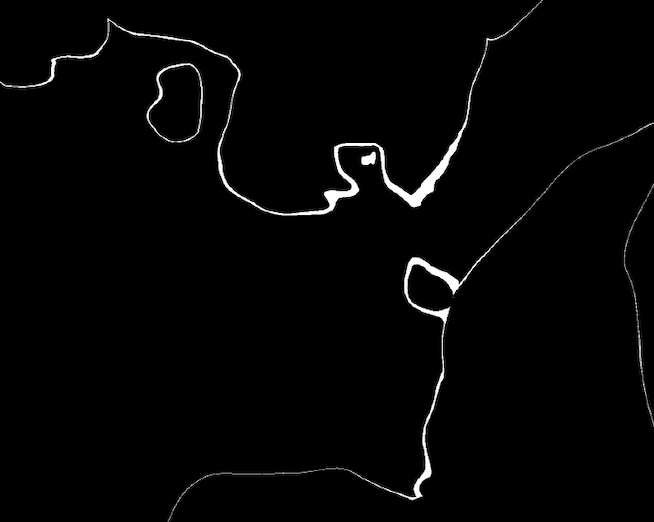}} 
    \subfigure[]{\label{fig:3c}\includegraphics[scale=0.5]{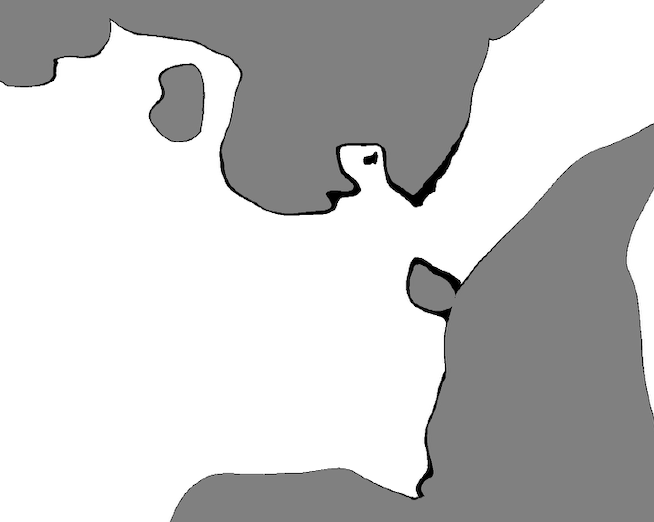}}
    \subfigure[]{\label{fig:3d}\includegraphics[scale=0.5]{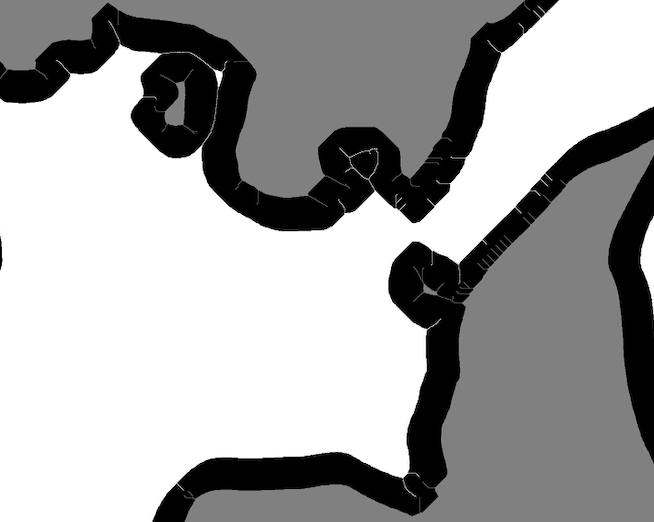}}    
    \subfigure[]{\label{fig:3e}\includegraphics[scale=0.5]{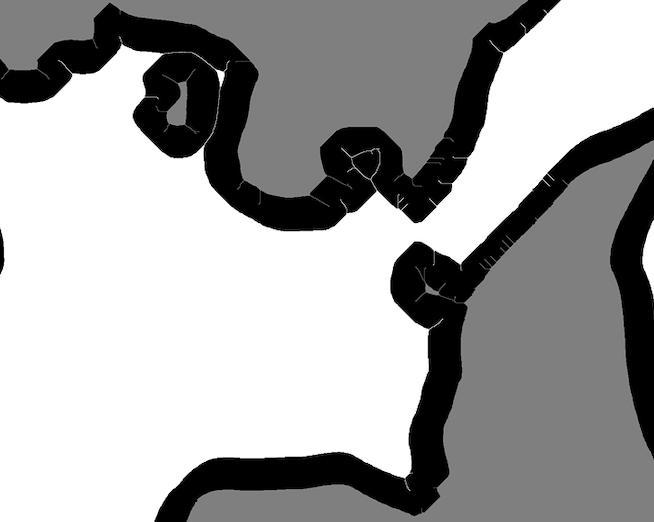}} \\    
  \caption{(a) a cropped portion of a high resolution image from BIG dataset \cite{cheng2020cascadepsp}. (b) Pixels for which the difference between bicubic interpolated label probabilities corresponding to the top two classes is below a threshold are highlighted in white. These are discarded from the potential seeds. (c) estimated labels by applying a softmax on the bicubic interpolated label probabilities on the remaining pixels. Grey corresponds to potential background seeds, white corresponds to potential object seeds and black corresponds to neither. (d) The potential seeds for both the object and background are further trimmed by applying a thinning iteratively. (e) final set of seeds obtained by discarding pixels by applying pruning iteratively.}
  \label{fig:MorphologicalUHDSemSegSeed}
\end{figure*}

As pixels closer to object boundaries are known to have unreliable semantic information, the potential seed set needs to be further trimmed by removing pixels closer to object boundaries. We decompose the potential seed set into disjoint subsets, one subset for each distinct label. Treating each such subset as foreground, we apply a standard morphological thinning on the foreground using the eight structuring elements \cite{gonzales2018digital}. The thinning is applied recursively for a preset number of iterations to trim the subsets of potential seeds for each distinct label (see Fig \ref{fig:3d}). To avoid branch-like structures that may degrade the performance of the random walker, we apply morphological pruning \cite{gonzales2018digital} to each of the subsets of potential seeds to discard a few more pixels (see Fig \ref{fig:3e}). The final set of seeds is then given by the union of outputs of all the pruned subsets. The reader may refer to Fig \ref{fig:MorphologicalUHDSemSegSeed} for a visual explanation on the details of seed selection block.

\paragraph{\textbf{Edge weights construction}} To construct edge weights, we apply Dollar gradient \cite{dollar2013structured} on the high resolution image. Dollar gradient estimates the probability that a pixel belongs to the boundary. Using the estimated probabilities, we set the edge weights of each edge $e_{ij}$ incident on the vertices $i$ and $j$ as follows:

\begin{equation}
\label{eq:EW}
w_{ij} = exp\left\{-\beta \frac{(p_i + p_j)^2}{\sigma} \right\}
\end{equation}

where $p_i$ (respectively $p_j$) is the probability that pixel $i$ (respectively $j$) is a boundary pixel, $\sigma$ is the standard deviation of the probabilities (of being a boundary pixel) of all pixels in the image, and $\beta$ is a hyperparameter. 

Intuitively is is clear that the higher the probability of pixels incident on edge $e_{ij}$, the lower the edge weight. Hence, the edge weights are large within objects and small across objects i.e. reflect a similarity measure between the pixels. The reader may refer to the image in Fig \ref{fig:1e} for visualizing the edge weights.

\paragraph{\textbf{Seed propagation method}} To propagate seeds, we apply the classic random walker algorithm used for image segmentation \cite{grady2006random}. Recall that the random walker can be cast as a constrained optimization problem. Random walker can be efficiently solved using a linear system of equations. This is done by rearranging the indices of all pixels such that all the seeds appear before the non-seeds. After rearrangement, one can decompose the Laplacian matrix $L$ of the image graph as follows:

\begin{equation}
L = \begin{pmatrix}
   L_{seed} & B\\\
B^T & L_U
 \end{pmatrix}
 \label{eq:LaplacianDecomp}
\end{equation}

The cost function of the random walker $RWCost(\textbf{x})$ can then be written as:

\begin{equation}
RWCost(\textbf{x}) = \frac{1}{2}(\textbf{x}_{seed}^TL_{seed}\textbf{x}_{seed} + 2\textbf{x}_U^TB^T\textbf{x}_{seed} + \textbf{x}_U^TL_U\textbf{x}_U),
\label{eq:RWCostDecomp}
\end{equation}

Here $\textbf{x}$ refers to the vector of label probabilities of all pixels, $\textbf{x}_U$, $\textbf{x}_{seed}$ denote the sub-vectors of $\textbf{x}$ corresponding to the unlabelled points and seeds respectively. The solution to random walker is the minimizer of Eq \ref{eq:RWCostDecomp} subject to $\textbf{x}_{seed}  = \textbf{f}_{seed}$ where $\textbf{f}_{seed}$ denotes the labels of the seeds. Applying elementary calculus, it can be seen that the solutions to the minimization problem can be obtained by solving the following linear system of equations:

\begin{equation}
L_U\textbf{x}_U = -B^T\textbf{x}_{seed}
\label{eq:RWCritical}
\end{equation}

Note that the above formulation corresponds to two-class image segmentation. The same ideas can be easily extended to multi-class image segmentation.

It is easy to see that our method can be trained without requiring the ground truth annotations of the high resolution images. This is because the seed selection only involves the usage of supervised models on the low resolution images. The edge weight construction requires only the high resolution input images.

\subsection{Relation to Existing Approaches}

In this subsection, we briefly discuss the connections between our pipeline and the existing approaches.

\paragraph{\textbf{Fusion of Local and Global Information}} Similar to CascadePSP \cite{cheng2020cascadepsp} and GLNet \cite{chen2019collaborative}, our pipeline can be viewed as a method that fuses local and global information. This is because the gradients obtained from the high resolution image corresponds to local information. On the other hand, the seed information for each of the semantic instances that we gather from the low resolution image can be considered as global information. This is because object information cannot be obtained by processing local windows separately and fusing the outputs.

\paragraph{\textbf{Non-Uniform Downsampling}} Non-uniform downsampling works by preserving the `uncertain' regions and downsampling the `certain' regions. Our approach to seed selection, in essence, has a similar effect.

\section{Experiments and Ablation Studies}
\label{sec:Experiments}

In subsection \ref{subsec:datasets}, the datasets used for the experiments are described. Then in subsection \ref{subsec:Comparison}, the results obtained by our pipeline are quantitatively and qualitatively compared to existing approaches. In subsection \ref{subsec:Robust}, the robustness of our method to the hyperparameters used in the seed selection block (see Fig \ref{fig:Schematic}) and robustness to the input learned gradient used in the edge weight construction block (see Fig \ref{fig:Schematic}) is established empirically. In subsection \ref{subsec:limitations}, the limitations of our pipeline are described in detail. 

\subsection{Datasets and Evaluation Measures}
\label{subsec:datasets}

For the experiments, we use two datasets namely BIG dataset \cite{cheng2020cascadepsp}, a high resolution image dataset and the relabelled PASCAL VOC 2012 \cite{cheng2020cascadepsp}. BIG dataset contains a total of $150$ high resolution images of sizes ranging from $2048 \times 1600$ to $5000 \times 3600$ with $20$ semantic classes (excluding the background). The BIG dataset does not contain any low resolution images. Hence, we downsample all the images in the BIG dataset such that the length of long axis is $512$ by maintaining the aspect ratio for the experiments in table \ref{tab:IoU}. The relabelled PASCAL VOC 2012 consists of $501$ images of sizes ranging from  $ 266 \times 400 $ to $459 \times 500$ with $20$ semantic classes (excluding the background). As the images in PASCAL VOC 2012 are not high resolution images, we apply our pipeline skipping the bi-cubic interpolation step in the seed selection process (see seed selection block in Fig \ref{fig:Schematic}) for the experiments. The experiments on the relabelled PASCAL VOC 2012 can thus be treated as refinement of the baseline methods. We use Eq \ref{eq:AvgIoU} (denoted as overall $\% IoU$) as the measure for evaluating the performance of semantic segmentation methods.

\subsection{Comparison to the Existing State-of-the-art}
\label{subsec:Comparison}

We perform both quantitative and qualitative comparison of our pipeline to existing approaches. For comparison we use baseline methods that work only on low resolution images i.e. FCN-8 \cite{long2015fully}, RefineNet \cite{lin2017refinenet}, PSPNet \cite{zhao2017pyramid}, and DeepLabV3+ \cite{chen2018encoder}, and refinements of baseline methods using the cascade approach \cite{cheng2020cascadepsp} developed for ultra high resolution segmentation, and MeticulousNet \cite{yang2020meticulous}, another ultra high resolution segmentation approach. The results of these methods are gathered from \cite{cheng2020cascadepsp,yang2020meticulous}. We implement our method using algorithms DeepLabV3+ and FCN-8 on the low resolution images for the seed selection process and Dollar gradient as the learned gradient on the high resolution counterpart to construct edge weights. Recall that the hyperparameters of our approach (described in subsection \ref{subsec:Core}) are threshold on the absolute difference between the top two probable classes for discarding not-so-sure pixels from seed set denoted by $t$, number of thinning iterations denoted by $n_{thin}$,  number of pruning iterations denoted by $n_{prun}$, and smoothing parameter $\beta$ in Eq \ref{eq:EW}. Table \ref{tab:HyperParameters} contains all the hyperparameters used for the experiments.

\begin{table}[tbhp]
\caption {Hyperparameters used for comparison with existing state-of-the-art}
\begin{tabular}{ |p{1.0cm}|p{1.25cm}|p{1.35cm}|p{1.25cm}|p{1.35cm}|  }
 \hline
\multicolumn{1}{|c|}{ } & \multicolumn{2}{|c|}{BIG} & \multicolumn{2}{|c|}{PASCAL VOC 2012} \\
 \hline
  &  FCN-8 + Ours  & DeepLabV3+ + Ours & FCN-8 + Ours  & DeepLabV3+ + Ours    \\
 \hline
  $t$ & 0.4 & 0.03 & 0.95 & 0.4\\ 
  $n_{thin}$ & 85 & 35 & 0 & 0\\   
  $n_{prun}$ & 10 & 20 & 0 & 0\\   
  $\beta$  & 10.8 & 11.3 & 10.9 & 10.5\\ 
 \hline
\end{tabular}
\label{tab:HyperParameters}
\end{table}

\paragraph{Quantitative Analysis}

Table \ref{tab:IoU} provides a summary of the performance of our method. On BIG dataset, our refinement of FCN-8 provides an $\%$IoU of $83.74$ which is higher than the overall $\%$IoUs of FCN-8, FCN-8 + Cascade, and MeticulousNet. A similar result holds for the refinement of DeepLabV3+. Also, our refinement of DeepLabV3+ are the best performing models with overall $\%$IoUs of $94.08$ and $93.16$ on BIG and PASCAL VOC 2012 respectively. 

\begin{table}[tbhp]
\caption {On both the datasets, w.r.t. overall $\%$IoU, our refinements of FCN-8 and DeepLabV3+ outperform the respective refinements using CascadeNet. Also, our refinement of DeepLabV3+ achieves the best results.}
\begin{tabular}{ |p{4cm}|p{1.5cm}|p{1.5cm}|  }
 \hline
 Method & BIG & PASCAL VOC 2012\\
 \hline
 FCN-8&   72.39   & 68.85\\
 RefineNet & 90.20 &  86.21\\ 
 PSPNet    &90.49 &  90.92\\
 DeepLabV3+   &  89.42  &   87.13\\ 
 \hline
 FCN-8 + Cascade &   77.87   & 72.70\\ 
 RefineNet + Cascade &92.79 &  87.48\\ 
 PSPNet + Cascade    &93.93 &  92.86\\ 
 DeepLabV3+ + Cascade  &  92.23  &   89.01\\  
 \hline
 MeticulousNet & 92.23 & NA \\
 \hline
 FCN-8 + Ours & 83.74 & 74.72\\ 
 DeepLabV3+ + Ours & \textbf{94.08}  & \textbf{93.16} \\
 \hline
\end{tabular}
\label{tab:IoU}
\end{table} 

\begin{figure*}[h]
    \subfigure[]{\label{fig:4a}\includegraphics[scale=0.035]{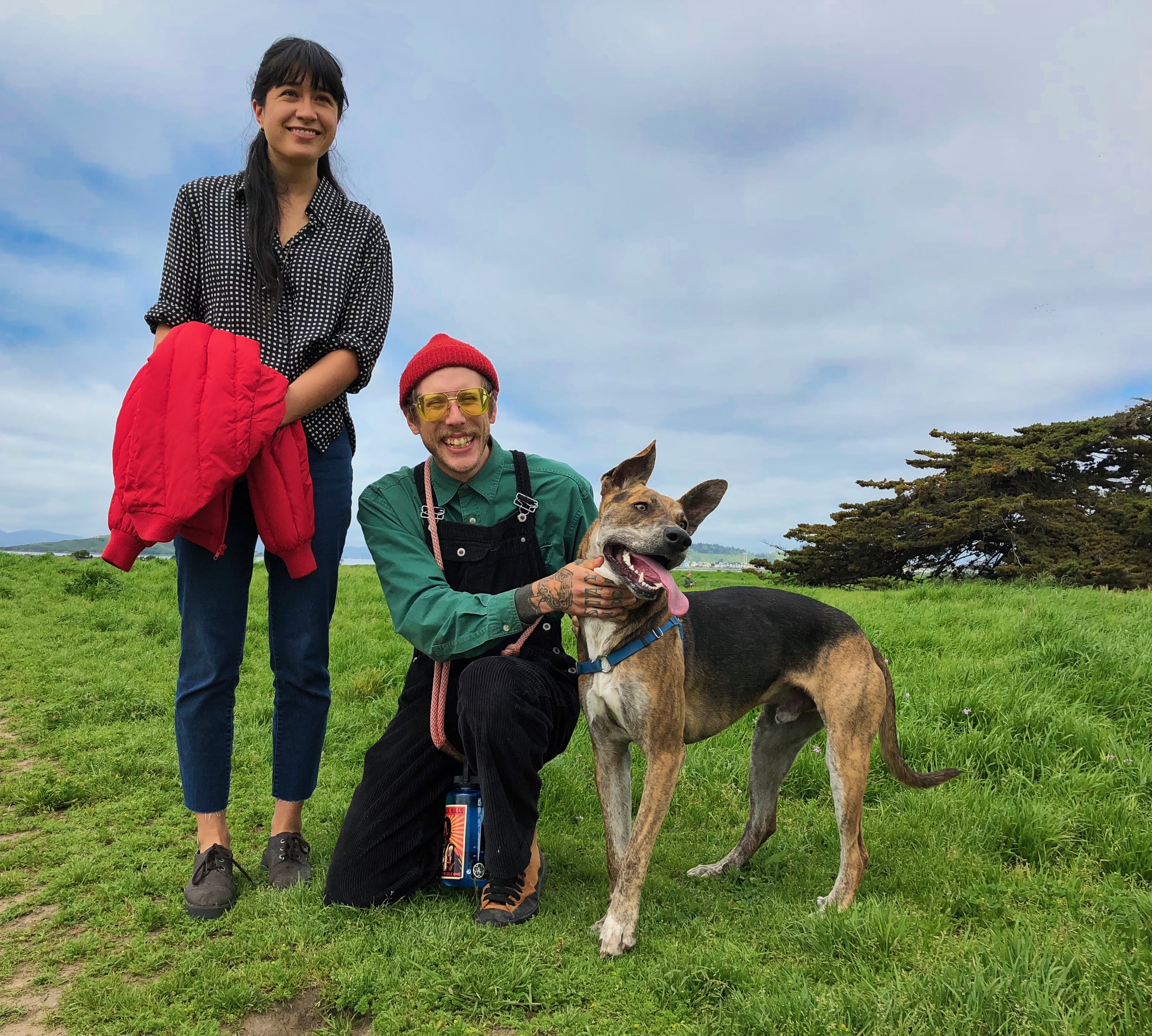}}
    \subfigure[]{\label{fig:4b}\includegraphics[scale=0.035]{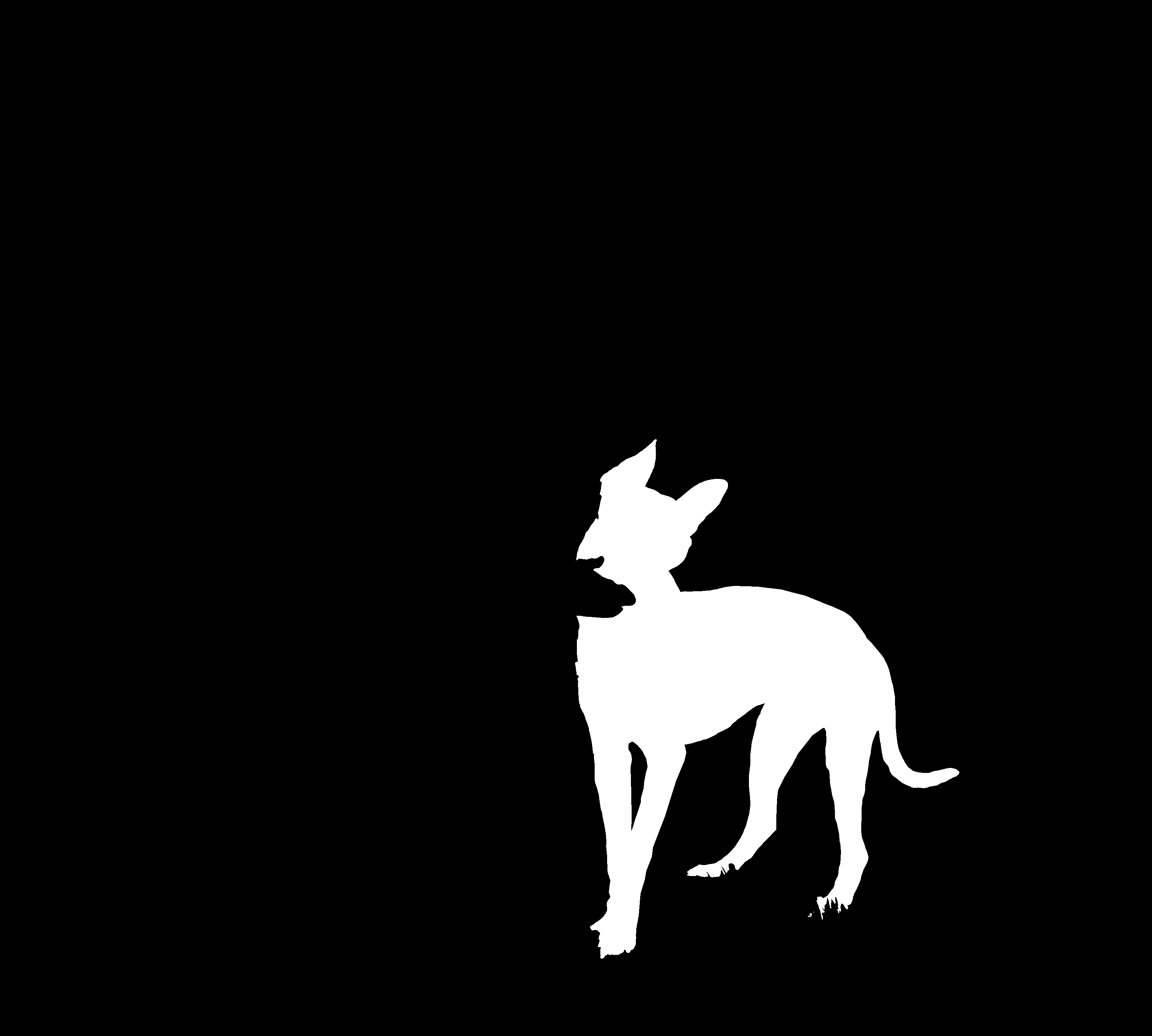}} 
    \subfigure[]{\label{fig:4c}\includegraphics[scale=0.035]{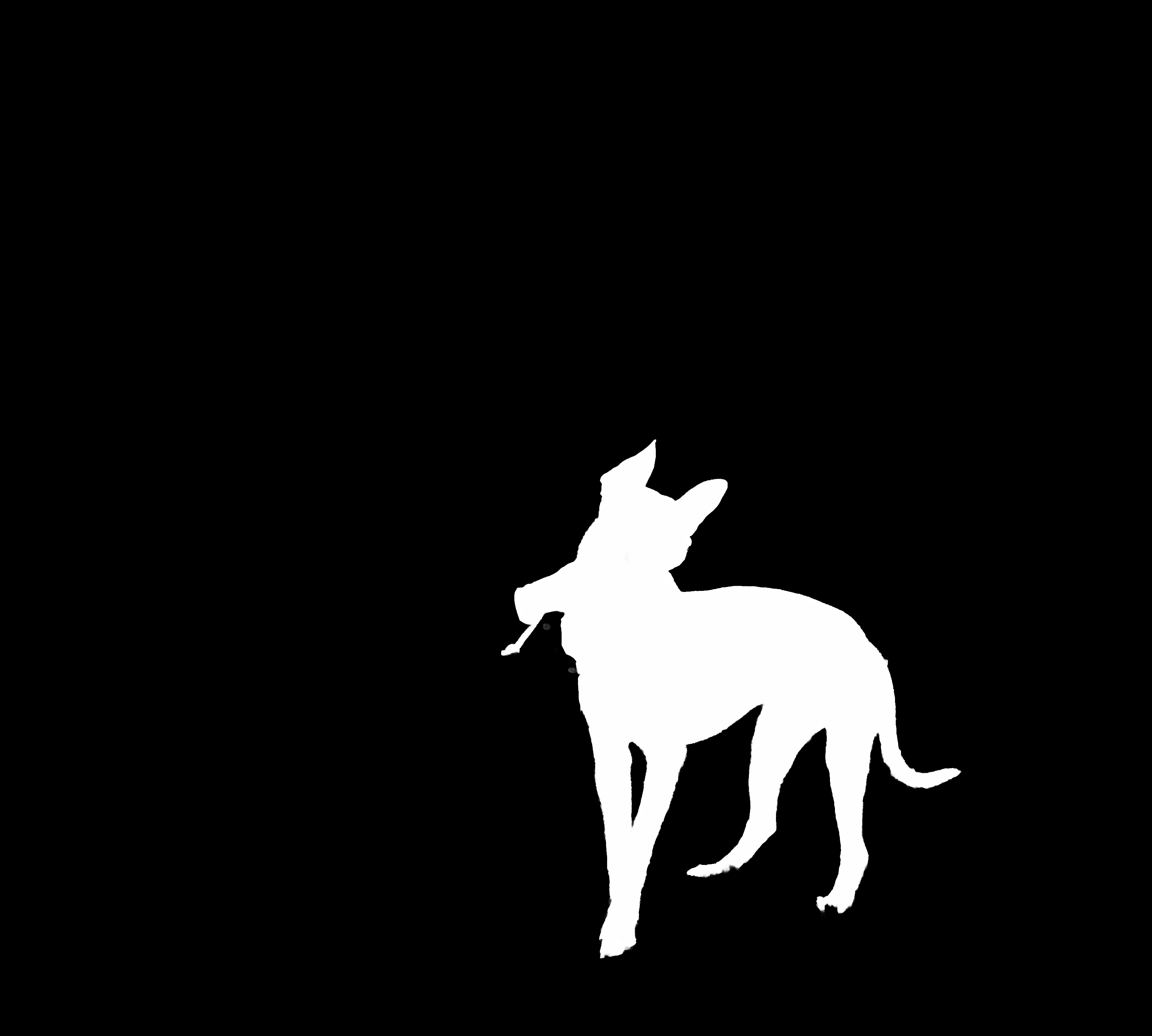}}
    \subfigure[]{\label{fig:4d}\includegraphics[scale=0.035]{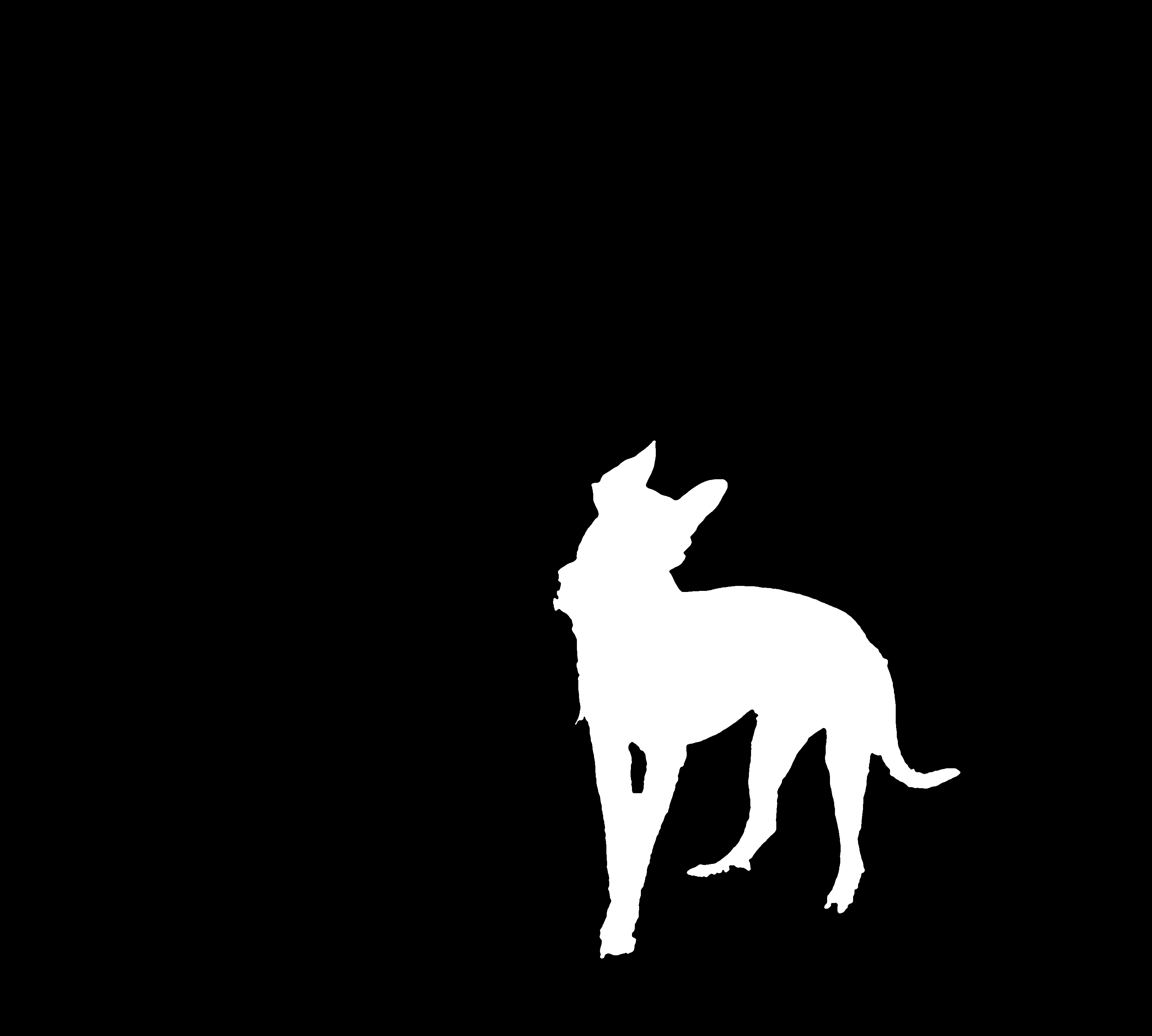}}    
    \subfigure[]{\label{fig:4e}\includegraphics[scale=0.045]{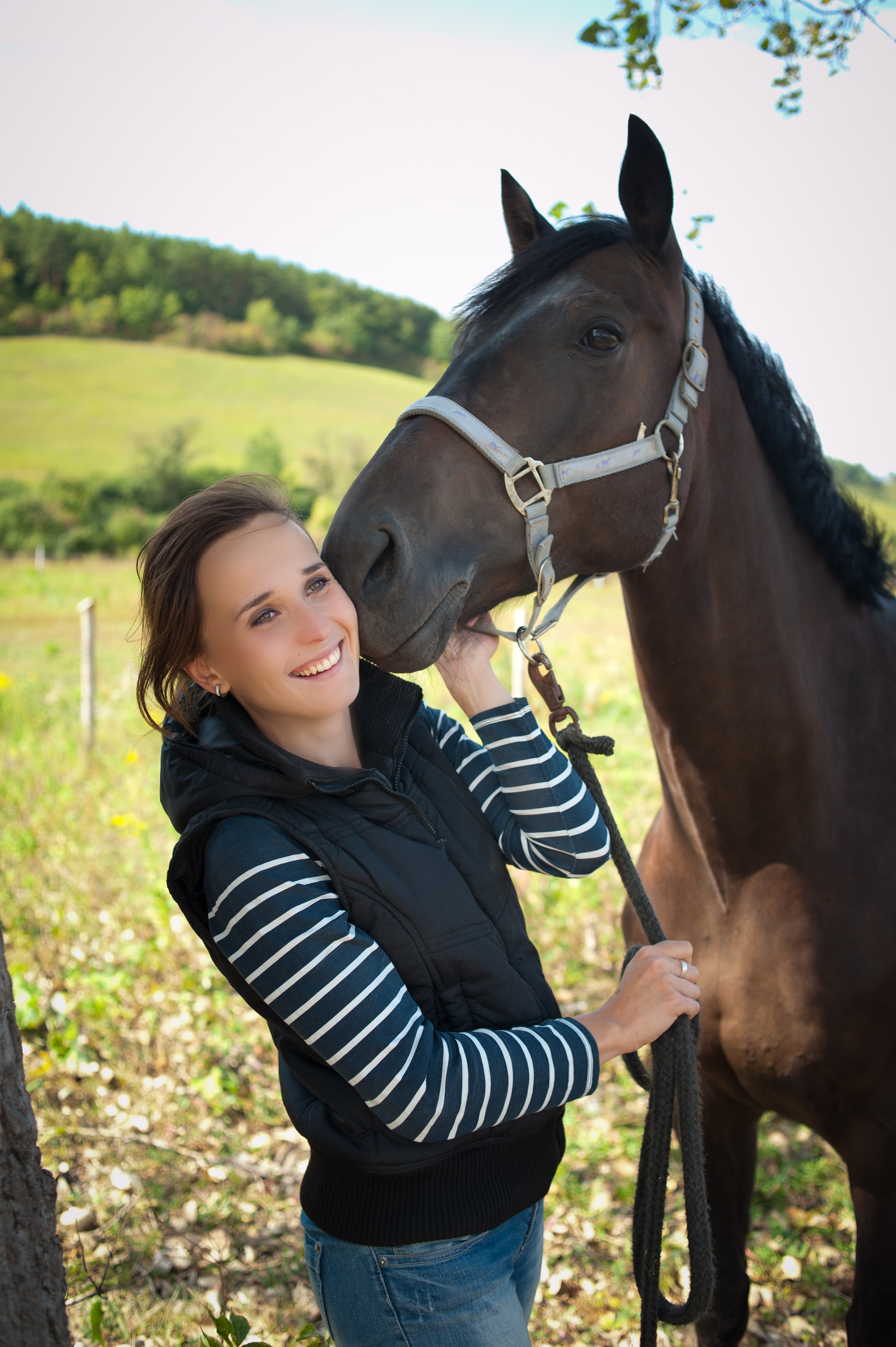}}
    \subfigure[]{\label{fig:4f}\includegraphics[scale=0.045]{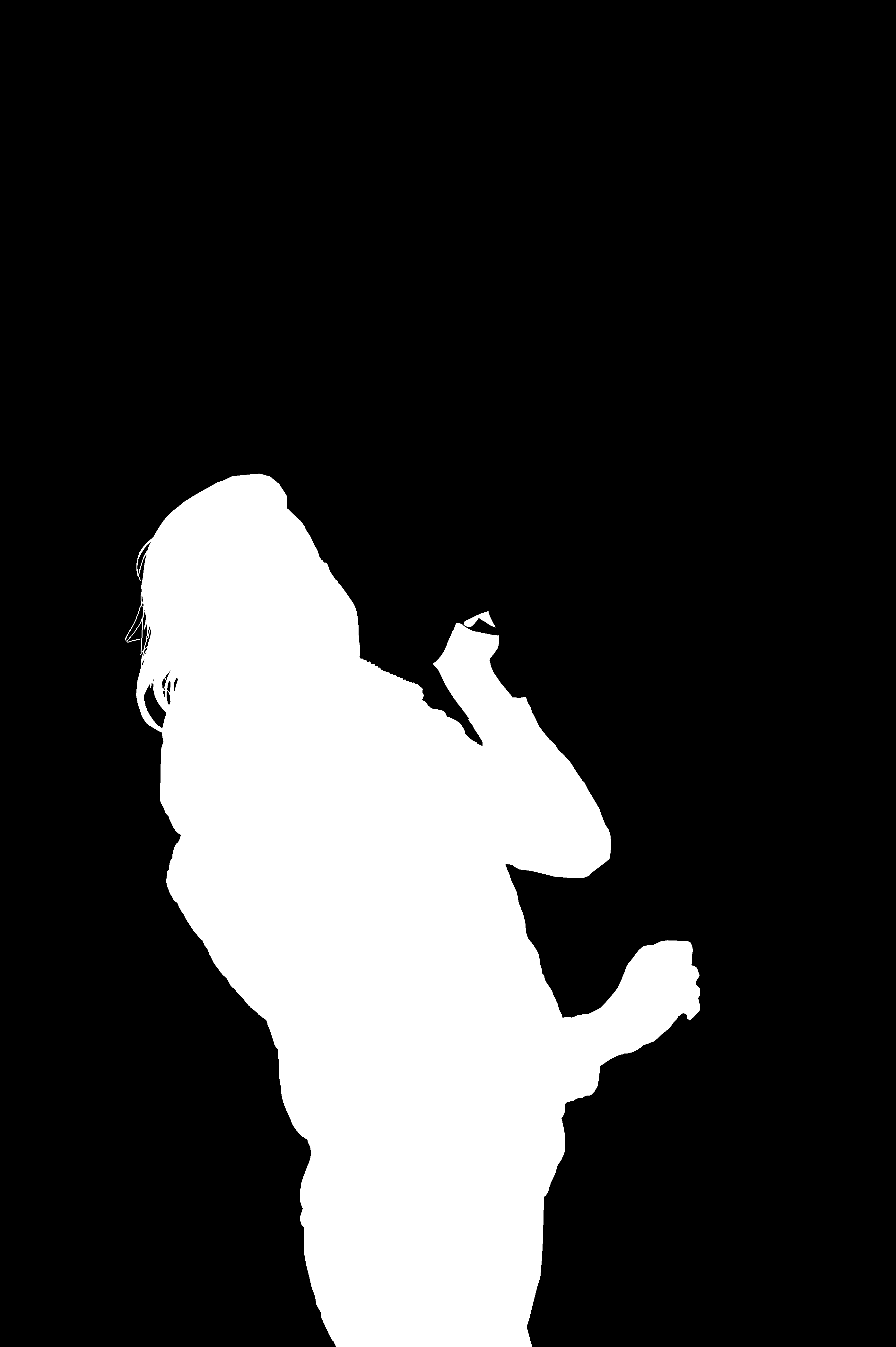}}     
    \subfigure[]{\label{fig:4g}\includegraphics[scale=0.045]{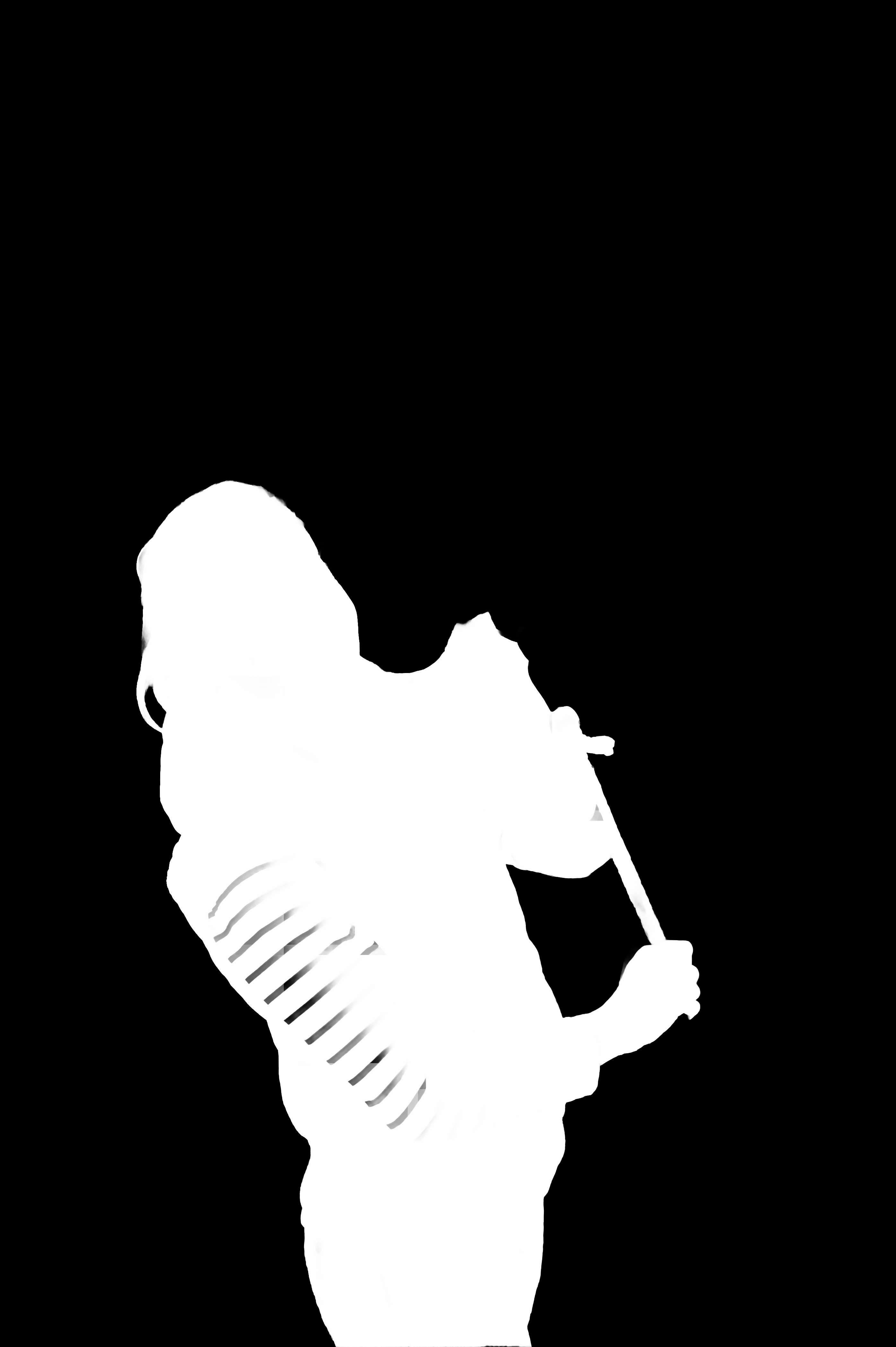}}     
    \subfigure[]{\label{fig:4h}\includegraphics[scale=0.045]{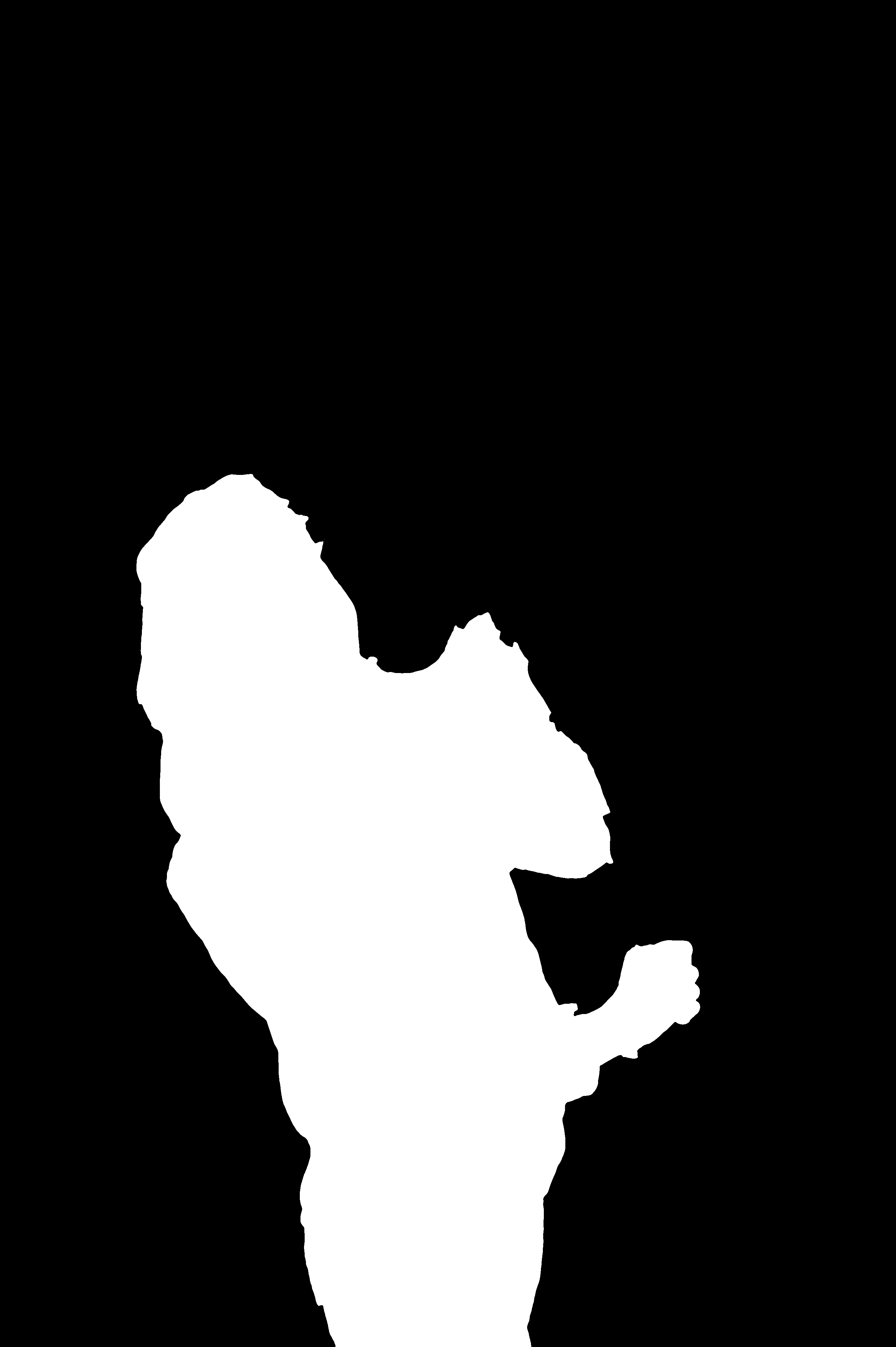}} \\    
  \caption{(a) and (e) Two images from the BIG dataset. (b) and (f) Ground truth semantic segmentation of both the images illustrating the classes dog and person respectively. (c) and (g) Results obtained by DeepLabV3+ + CascadeNet \cite{cheng2020cascadepsp}. (d) and (h) The results obtained by our pipeline using DeepLabV3+ as the base method on the corresponding low-resolution image and Dollar gradient for constructing edge weights. In (c), the hand is misclassified as dog. On the other hand, in (d) the portion of the hand not on dog is correctly classified. In (g), the stripes and the belt are incorrectly classified. While in (h), these are correctly classified. However, in (d), the portion of hand on the dog, and in (h), the pixels between left arm and left bicep are incorrectly classified. This is because our method cannot reconstruct objects with low isoperimetric quotient very well.}
  \label{fig:QualitativeComparison}
\end{figure*}

\paragraph{Qualitative Analysis}
Next, in Fig \ref{fig:QualitativeComparison} we provide a qualitative analysis based on visual comparison of the results of our method with the current state-of-the-art model namely CascadeNet. Two images are chosen from BIG for this purpose. The first column contains the images and the second column contains the ground truth semantic segmentations. The third column consists of the results obtained by DeepLabV3+ + CascadeNet. Notice that for the image in the first row, the hand is misclassified as dog. For the image in the second row, some of the striped patterns on the person are identified as not-person. Also, the belt is incorrectly identified as person. In the fourth column, the results from our method starting with DeepLabV3+ on low resolution images for seed selection are shown. Notice that the portion of hand that is not on the dog is classified correctly. However, the remaining portion of the hand is misclassified. Similarly, on the second image, our method performs better than Cascade refinement as the belt and stripes are classified correctly. However, our method misclassified the background pixels between the left arm and the left bicep. In fact, our method tends to slightly over-smooth the boundaries. More details on the limitations of our method are discussed in subsection \ref{subsec:limitations}. 

\subsection{Robustness Analysis}
\label{subsec:Robust}
In this subsection, we discuss the robustness of our method to the inputs of the algorithm. More specifically, the robustness to the hyper parameters used for seed selection and to the choice of learned gradient used to construct edge weights are discussed. 

\paragraph{Seed Quality}

\begin{figure}[h]
	\centering\	
	\includegraphics[width=0.95\columnwidth]{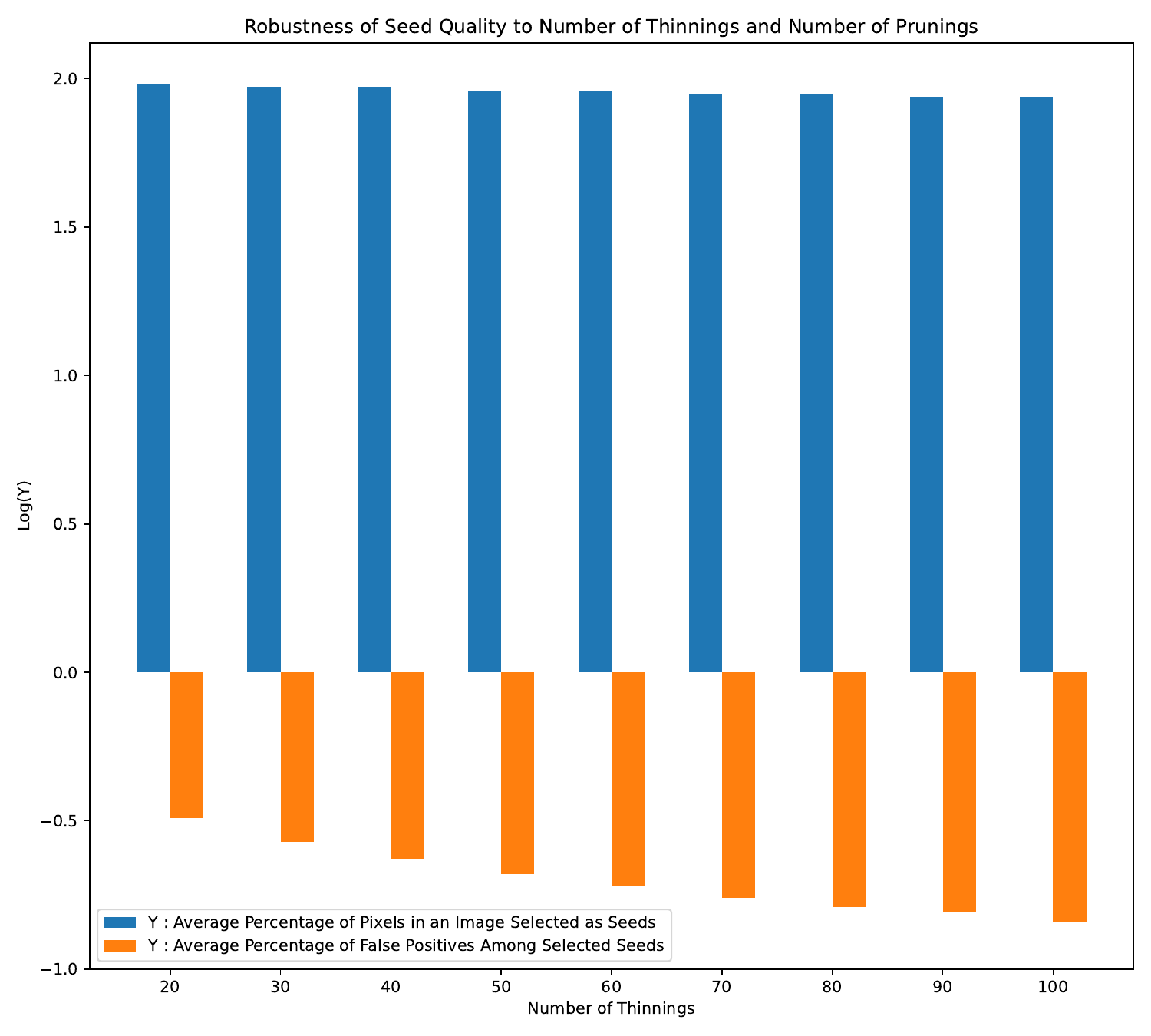} 											
	\caption{Average percentage of pixels chosen as seeds (blue bars), and average percentage of false positive seeds (orange bars) are plotted by varying the number of thinning operations in our pipeline on the BIG dataset. Percentages are plotted on a logarithmic scale for better visualization. As the number of thinning iterations are increased from $20$ to $100$, the pixels selected as seeds ranged from $97\%$ to $88\%$, and the number of false positive seeds ranged from $0.3\%$ to $0.1\%$. At each number of thinning iterations, the results were constant when the number of prunings were varied. This indicates robustness to the number of iterations of both thinning and pruning.}
\label{fig:ThinningPruningSeedQuality}
\end{figure}

The results from Fig \ref{fig:UpscaleErrorsAndBoundaries} imply that the pixels that are far away from boundaries are reliable as seeds. As the morphological operators we use for seed selection namely thinning and pruning are precisely meant for discarding boundary pixels, theoretically it is clear that the seeds selected from our pipeline would not contain many false positives i.e. any pixel selected has a seed for a particular semantic class has the same semantic ground truth label. We perform an experiment to check - 1) the average percentage of pixels in the image that we select as seeds, and 2) among the pixels that are selected as seeds, the average percentage of pixels that are selected as a seed for an incorrect class i.e. a false positive seed. These measures are computed for all classes together across all images in BIG, by varying the number of thinning iterations and the number of pruning iterations. A very high percentage of pixels selected as seeds and a very low percentage of false positives would imply that there are enough number of seeds for reconstructing the objects and that the quality of seeds are good. 

The number of thinning iterations are varied from $20-100$ with a step size of $10$ i.e. $9$ levels altogether. For each number of thinning iterations, number of pruning iterations are varied from $10-30$ with a step size of $5$ i.e. $5$ levels altogether. For a fixed number of thinning iterations, irrespective of the number of pruning iterations, it is observed that both these percentages are constant upto two decimals. For various number of thinning iterations, the percentage of pixels selected as seeds varied from $97\%$ to $88\%$ and the percentage of false positive seeds varied from $0.3\%$ to $0.1\%$ with the respective lower percentages corresponding to higher number of thinning iterations. As the number of thinning iterations are increased, the selection of seed pixels is more conservative as we tend to discard pixels slightly farther from object boundaries selected. Since the percentages are skewed, we plot the logarithm of the percentages (to the base $10$) in Fig \ref{fig:ThinningPruningSeedQuality} for better visualization. The blue bars correspond to the percentage of pixels selected as seeds on the logarithmic scale. The orange bars correspond to the percentage of selected seeds chosen for an incorrect class on the logarithmic scale. Observe that the orange bars are all negative indicating that these percentages are below $1\%$. The blue bars are all close to $2$ indicating that the percentages are very high (above $88\%$ to be precise). Hence the seed selection is robust to choice of number of thinning and number of pruning operations chosen in the pipeline.




\paragraph{Gradient Quality}

\begin{table}[tbhp]
\caption {Robustness to the noise in the learned gradient}
\begin{tabular}{ |p{0.85cm}|p{0.85cm}|p{0.85cm}|p{0.85cm}|p{0.85cm}|p{0.85cm}|p{0.85cm}|  }
 \hline
\multicolumn{1}{|c|}{ } & \multicolumn{3}{|c|}{Estimated} & \multicolumn{3}{|c|}{GT Refinement} \\
 \hline
 Noise Var - $\sigma^2$ & $\%$IoU - All  & $\%$IoU - Non-seed & $\frac{||\delta \textbf{x} ||}{||\textbf{x}||}$ - All &  $\%$IoU - All  &  $\%$IoU - Non-seed  &  $\frac{||\delta \textbf{x} ||}{||\textbf{x}||}$ - All \\
 \hline
0.02 & 94.10 & 75.55 & 0.04 & 98.36 & 88.45 & 0.04 \\ 
  0.1 & 94.05 & 75.21 & 0.13 & 98.32 & 88.16 & 0.13 \\ 
  0.2 & 93.96 & 74.50 & 0.20 & 98.25 & 87.67 & 0.19 \\ 
  0.4 & 93.79 & 73.25 & 0.26 & 98.07 & 86.49 & 0.25 \\ 
  0.5 & 93.71 & 72.66 & 0.28 & 97.99 & 85.95 & 0.27 \\ 
  0.6 & 93.62 & 72.02 & 0.30 & 97.89 & 85.30 & 0.28 \\ 
  0.8 & 93.38 & 70.35 & 0.34 & 97.53 & 83.03 & 0.33 \\ 
     1 & 93.10 & 68.40 & 0.38 & 97.14 & 80.56 & 0.36 \\    
 \hline
\end{tabular}
\label{tab:RobustnessLearnedGrad}
\end{table}

Another input to our pipeline is the learned gradient used to construct the edge weights. Recall that the solutions to a random walker are obtained by solving Eq \ref{eq:RWCritical}. Thus, the robustness of the solutions to edge weights depend on the Laplacian matrix given by Eq \ref{eq:LaplacianDecomp}. 

In order to empirically verify the robustness of our pipeline to the choice of the learned gradient, we perturb the probability map obtained by Dollar gradient with Gaussian noise $N(0,\sigma^2)$ at several levels of $\sigma^2$ (see table \ref{tab:RobustnessLearnedGrad}). The perturbed probabilities are then truncated with the lowest set to $0$ and highest to $1$. From the perturbed probabilities, edge weights are then constructed using Eq \ref{eq:EW}. We construct seeds in two different ways: 

\begin{enumerate}
\item We start with DeepLabV3+ on low resolution images (with long-axis of length $512$) of the BIG dataset to obtain low resolution estimates and eventually the seeds (see seed selection block in Fig \ref{fig:Schematic}). 
\item We start with the ground truth of BIG images and apply thinning and pruning to obtain seeds skipping the bicubic interpolation and difference threshold steps (see seed selection block in Fig \ref{fig:Schematic}). 
\end{enumerate}

The linear systems in Eq \ref{eq:RWCritical} are solved to obtain estimated label probabilities on both the perturbed and the original graphs (corresponding to each of the seed construction procedures). The other hyperparameters are same as described for the experiments on the BIG dataset in subsection \ref{subsec:Comparison}. We compute the overall $\%$IoU of estimated labels on the perturbed system, the $\%$IoU of estimated labels of only the unseeded pixels of the perturbed system, and the ratio of the norm of the difference of the estimated label probabilities (i.e. $\delta \textbf{x} = \textbf{x}_{perturbed} - \textbf{x}$) to the norm of label probabilities computed on the unperturbed graph ($\textbf{x}$). Note that at each level of noise variance, the results summarized in table \ref{tab:RobustnessLearnedGrad} are averages over $50$ iterations of random Gaussian noise. 

Firstly note that the $\%$IoUs (overall and non-seeds only) of the second experiment are consistently higher than that of the first experiment. This is expected as the second experiment uses ground truth for seed selection while the first uses estimates. For both the experiments, the overall $\%$IoU is reasonably good even at high levels of noise. This can be explained from the experiment on seed quality (see Fig \ref{fig:ThinningPruningSeedQuality}) which indicate that about $90\%$ of the pixels are selected as seeds and that the false positives are under $0.3\%$.

On the other hand, the quality of the gradient affects the $\%$IoU of the unseeded pixels. The columns on the relative ratio of the norm of the difference in estimated probabilities to the norm of the estimated probability of the unperturbed graph indicate that the gradients are not so informative for $\sigma^2 \geq 0.1$. This is because the ratio is computed on the estimates of all pixels together and a ratio of $0.13$ is a high number given that $\approx 90\%$ are correctly labelled from the seed selection. However, the $\%$IoU of non-seeds is steadily declining with increase w.r.t. the levels of noise variance. This implies that the quality of the results from our pipeline does not diminish abruptly with slight noise. Thus, our pipeline is robust to the choice of learned gradient used. 


\subsection{Limitations of Our Pipeline}
\label{subsec:limitations}

\paragraph{Scaling Factor between the Low Resolution and High Resolution Counterparts}
It is easily seen that if the scaling factor between the low resolution and high resolution image is too large, the quality of the results would diminish. This is obvious from the fact that the low resolution counterpart would completely miss some objects that are present in the high resolution image. In the next experiment, we study the performance of our model w.r.t. the scaling factor between the low and high-resolution images. For this experiment, we tested how the performance of our method changes as we down sample all the images in BIG dataset such that the length of the longer axis is $30, 90, 150, 210, 270, 330, 390, 450$ and $512$ while maintaining the aspect ratio. We then upsample them to the scale $512$ on the long-axis to pass it through DeepLabV3+ to begin the seed selection procedure (i.e. NN in Fig \ref{fig:Schematic}). The remaining steps in the pipeline are exactly as earlier. We observed that the average $\%$IoU is reasonably good when the low resolution image has a longer axis of length $150$ or more. In other words, since the dimensions of the smallest image in BIG is $2048 \times 1600$, one can conclude that our method works well upto a scaling factor of $13$ (see table \ref{tab:ScaleFactorIoU}). 

\begin{table}[tbhp]
\caption {Highest $\%$IoU achieved at each scale. Observe that the overall $\%$IoU is reasonably good when the length of longer axis is $150$ or more.}
\begin{tabular}{ |p{3.5cm}|p{3.5cm}|  }
 \hline
 Scale (Long axis length) & $\%$ IoU \\
 \hline
 30&  26.92    \\
 90 &  76.48\\ 
 150    &  89.99\\
 210   & 92.08\\ 
 270   & 93.03\\  
 330&    93.42  \\
 390 &  93.95\\ 
 450   & 94.16\\ 
 512   & 94.08\\   
 \hline
\end{tabular}
\label{tab:ScaleFactorIoU}
\end{table} 

\begin{figure*}[h]
    \subfigure[]{\label{fig:6a}\includegraphics[scale=0.03]{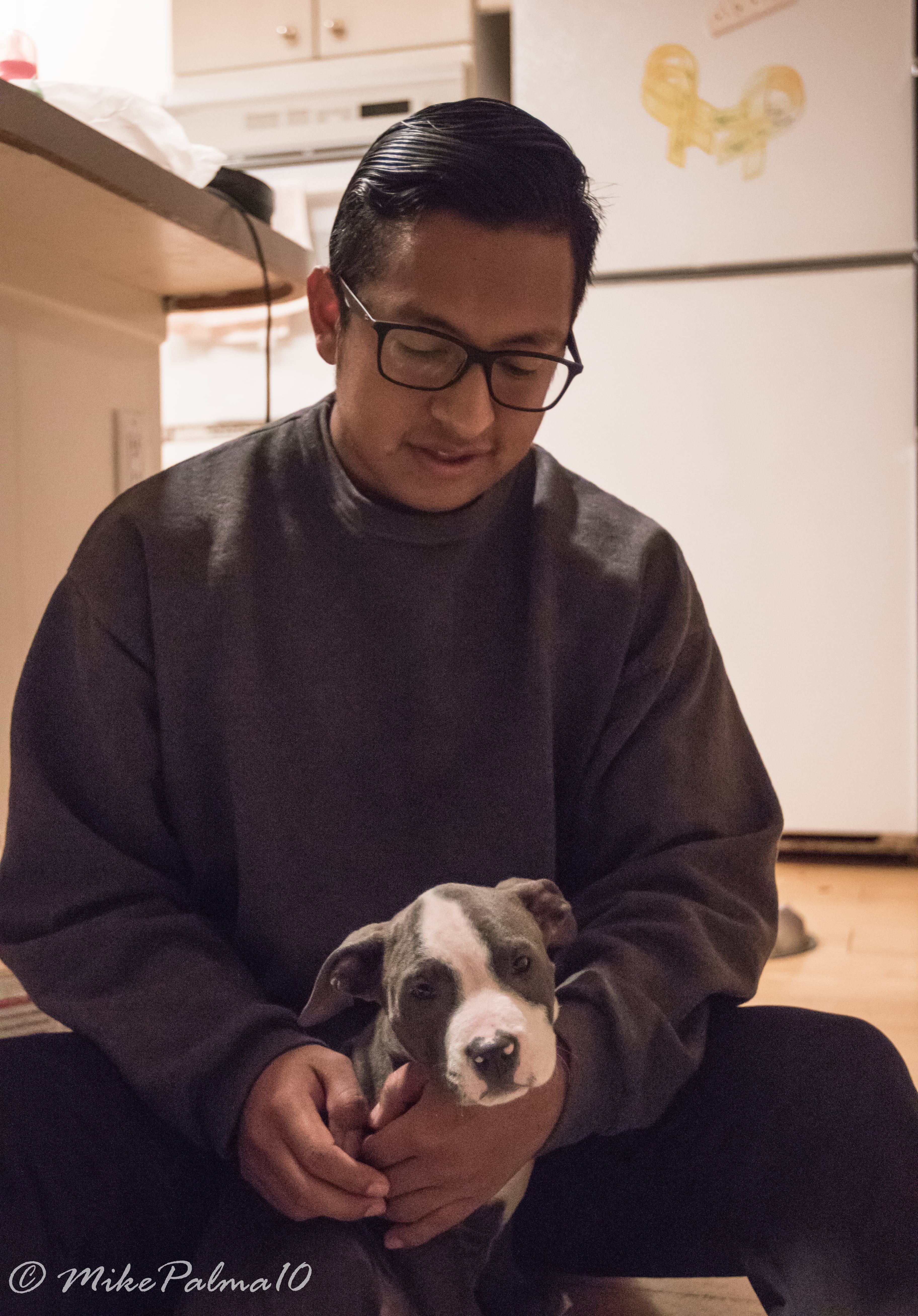}}
    \hspace{0.4cm}
    \subfigure[]{\label{fig:6b}\includegraphics[scale=0.03]{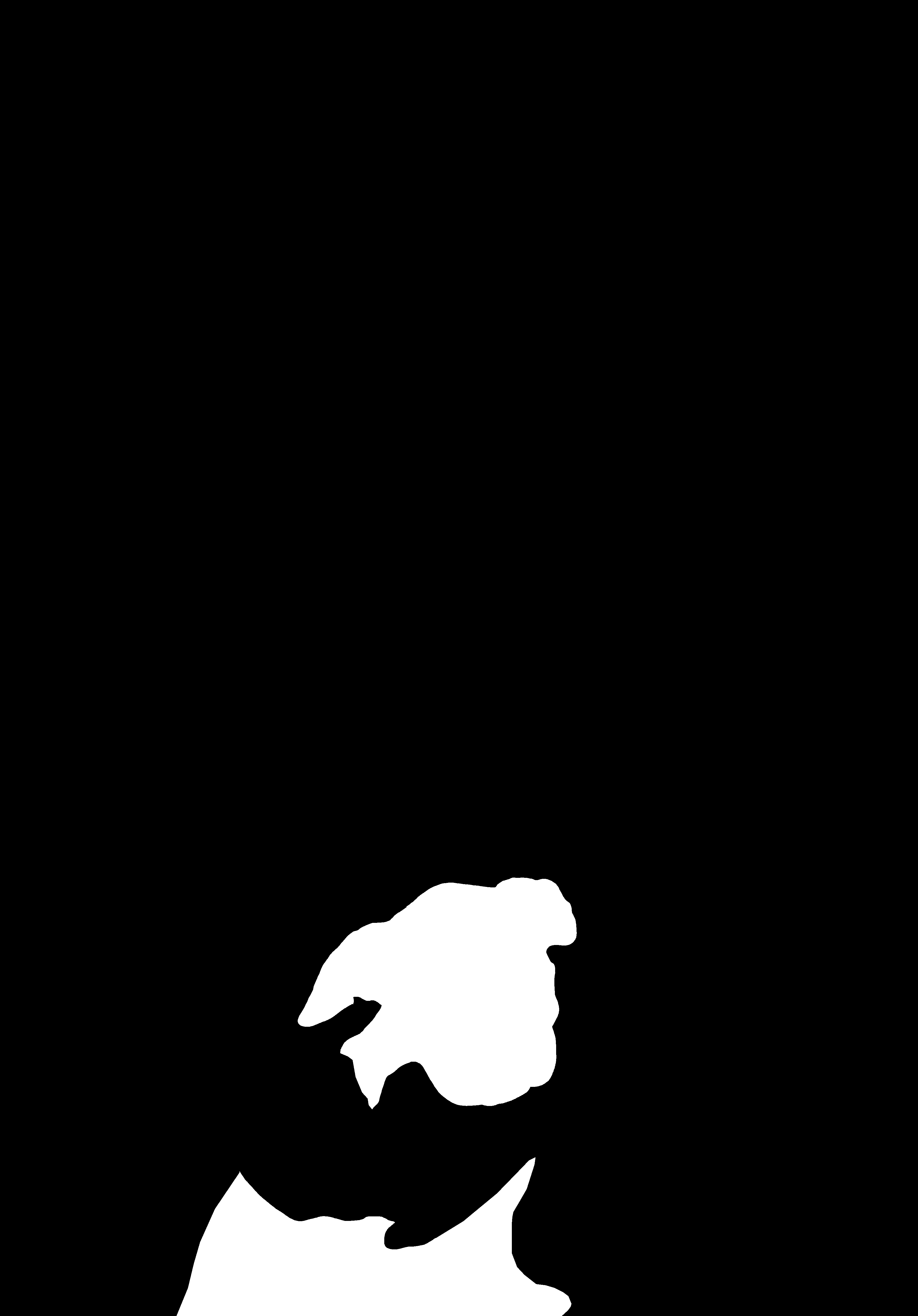}} 
    \hspace{0.4cm}
    \subfigure[]{\label{fig:6c}\includegraphics[scale=0.03]{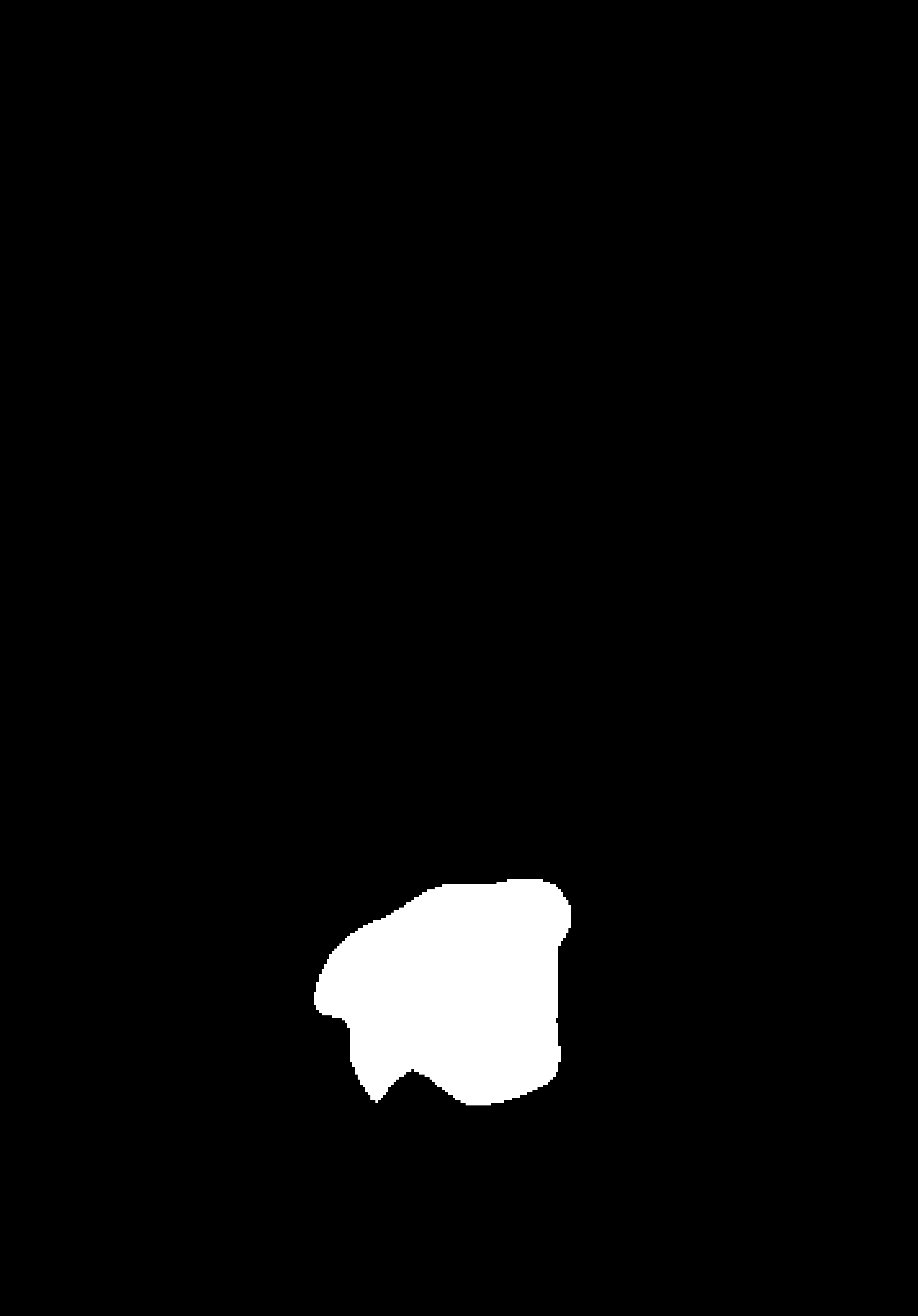}}
    \hspace{0.4cm}
    \subfigure[]{\label{fig:6d}\includegraphics[scale=0.03]{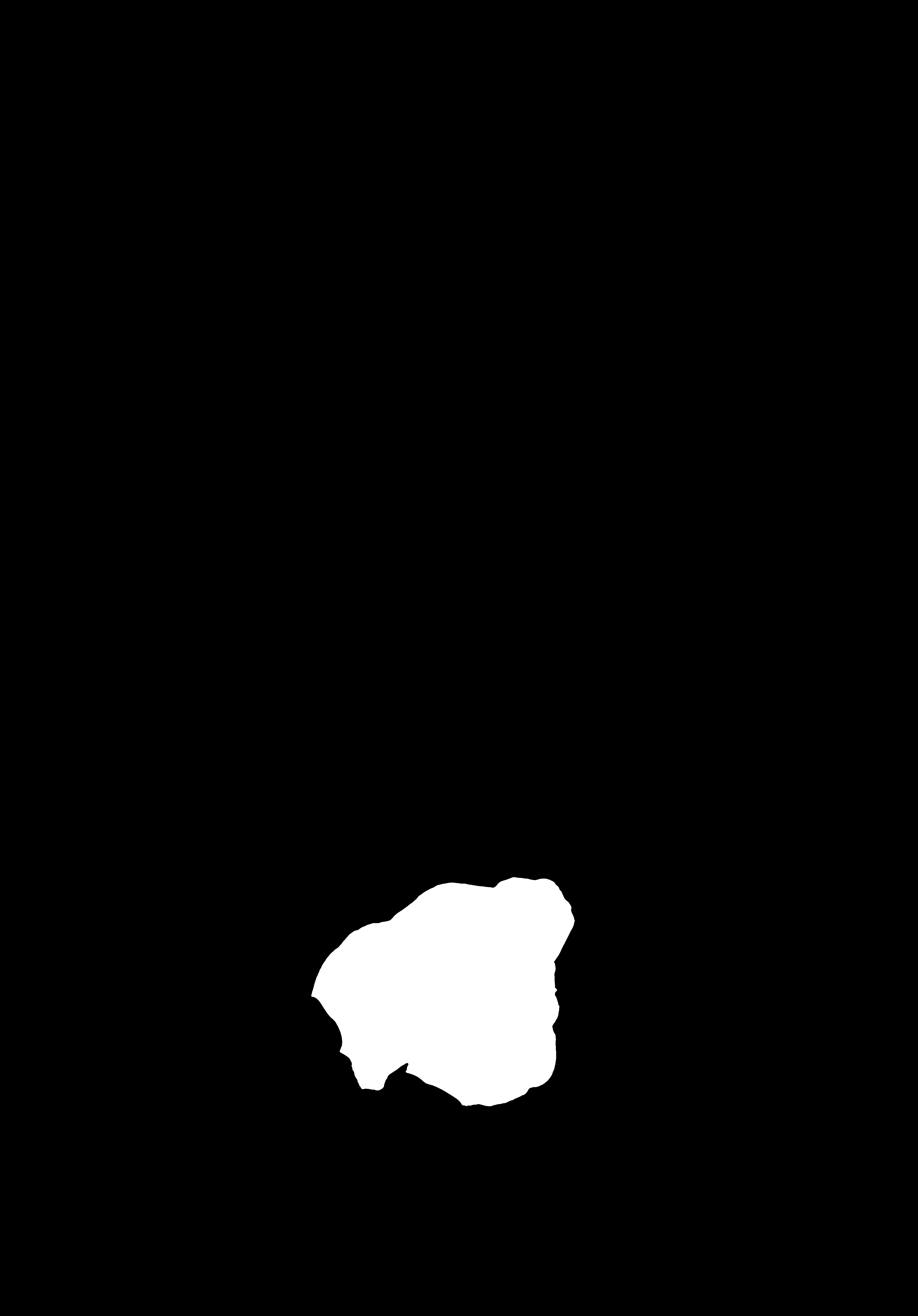}}    \\
    \subfigure[]{\label{fig:6e}\includegraphics[scale=0.03]{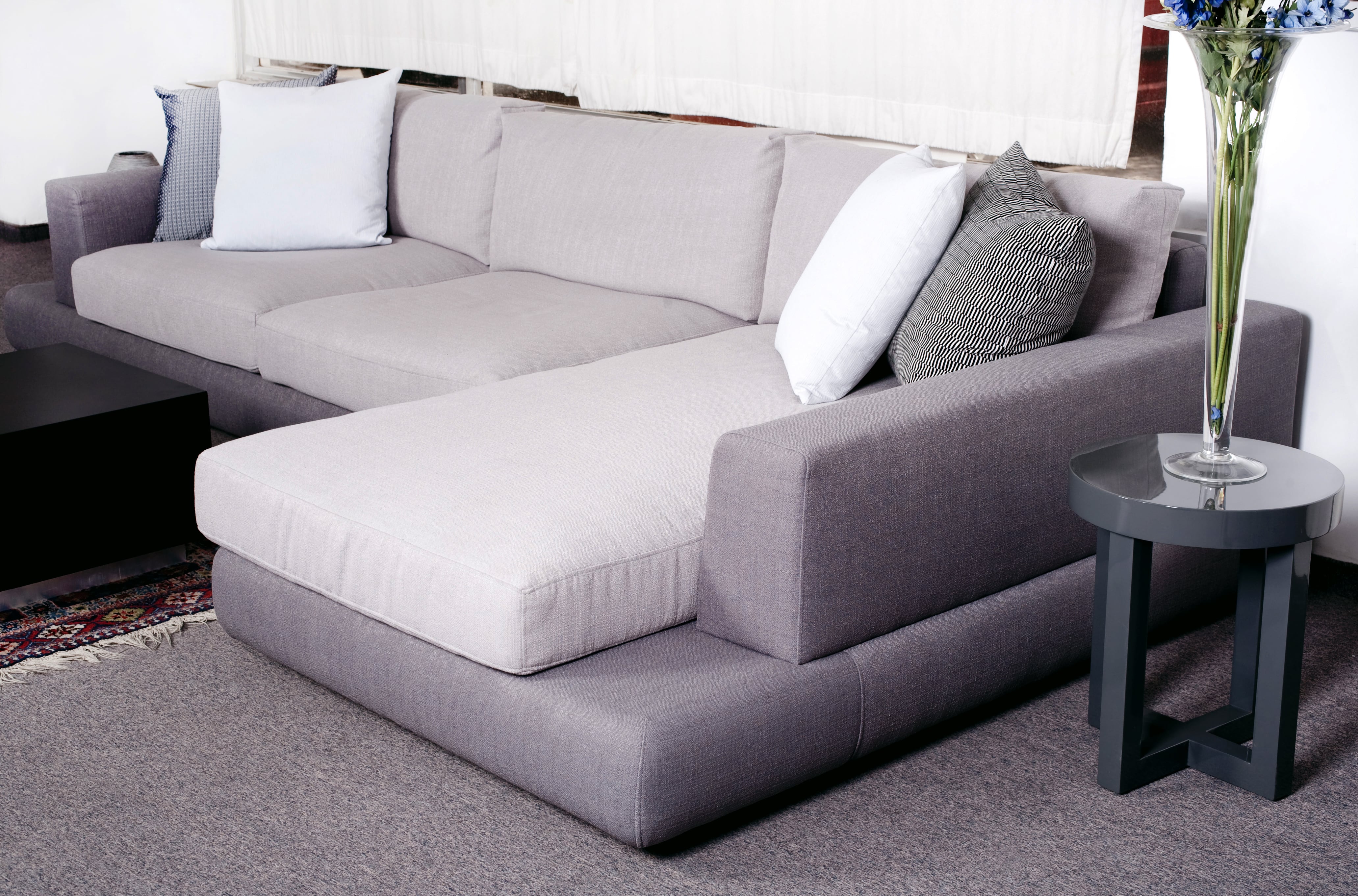}}
    \subfigure[]{\label{fig:6f}\includegraphics[scale=0.03]{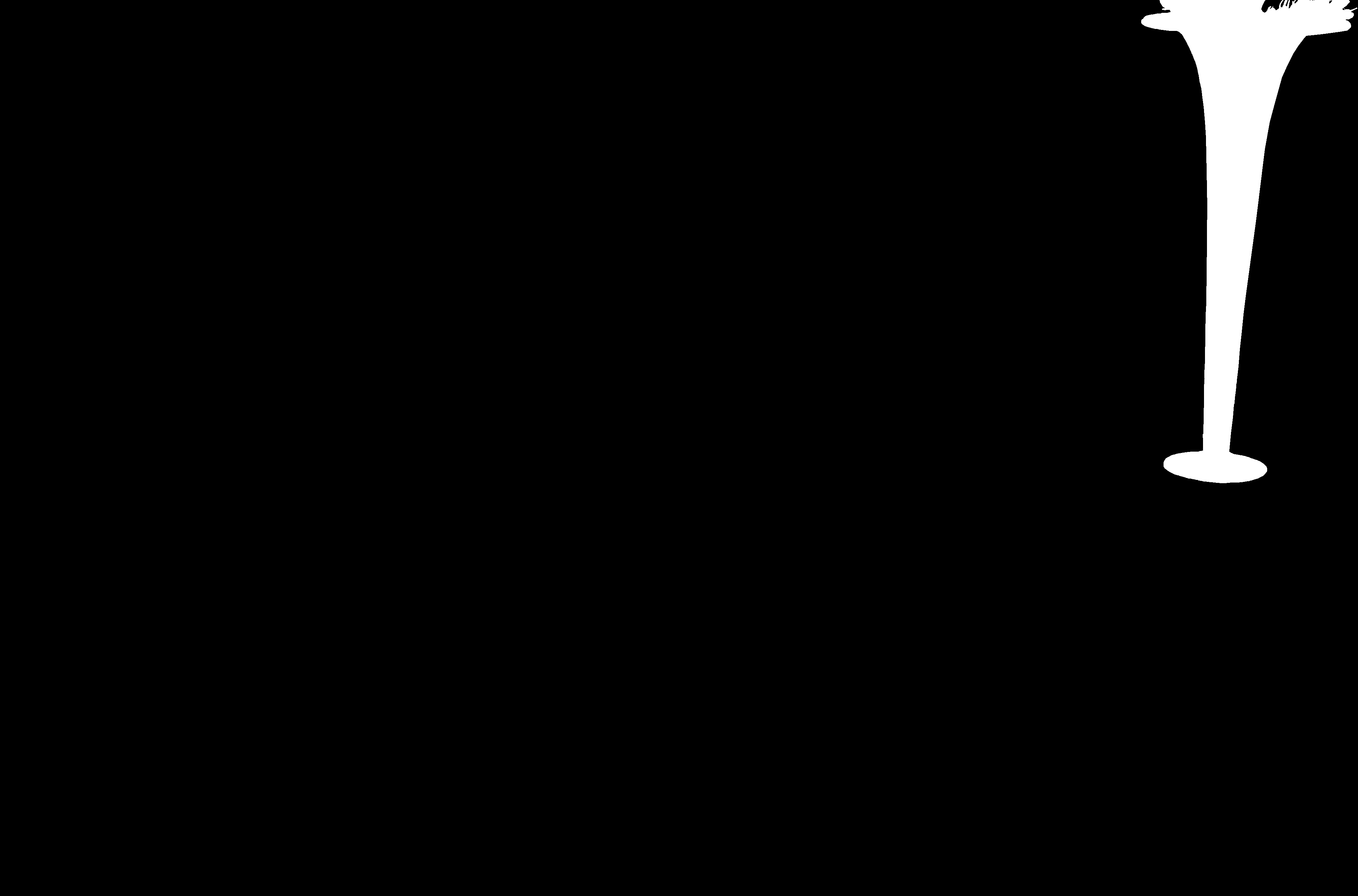}}     
    \subfigure[]{\label{fig:6g}\includegraphics[scale=0.03]{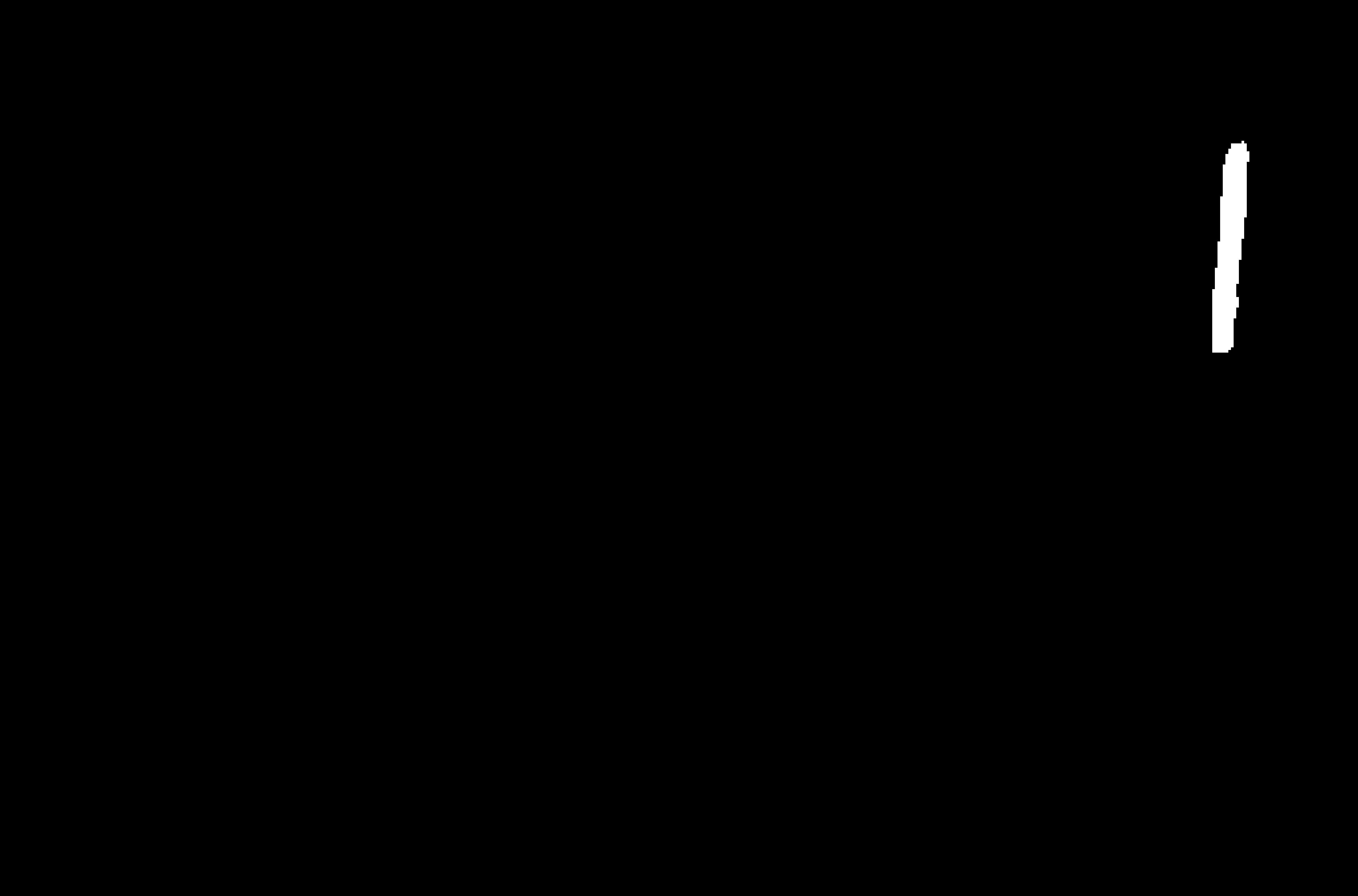}}     
    \subfigure[]{\label{fig:6h}\includegraphics[scale=0.03]{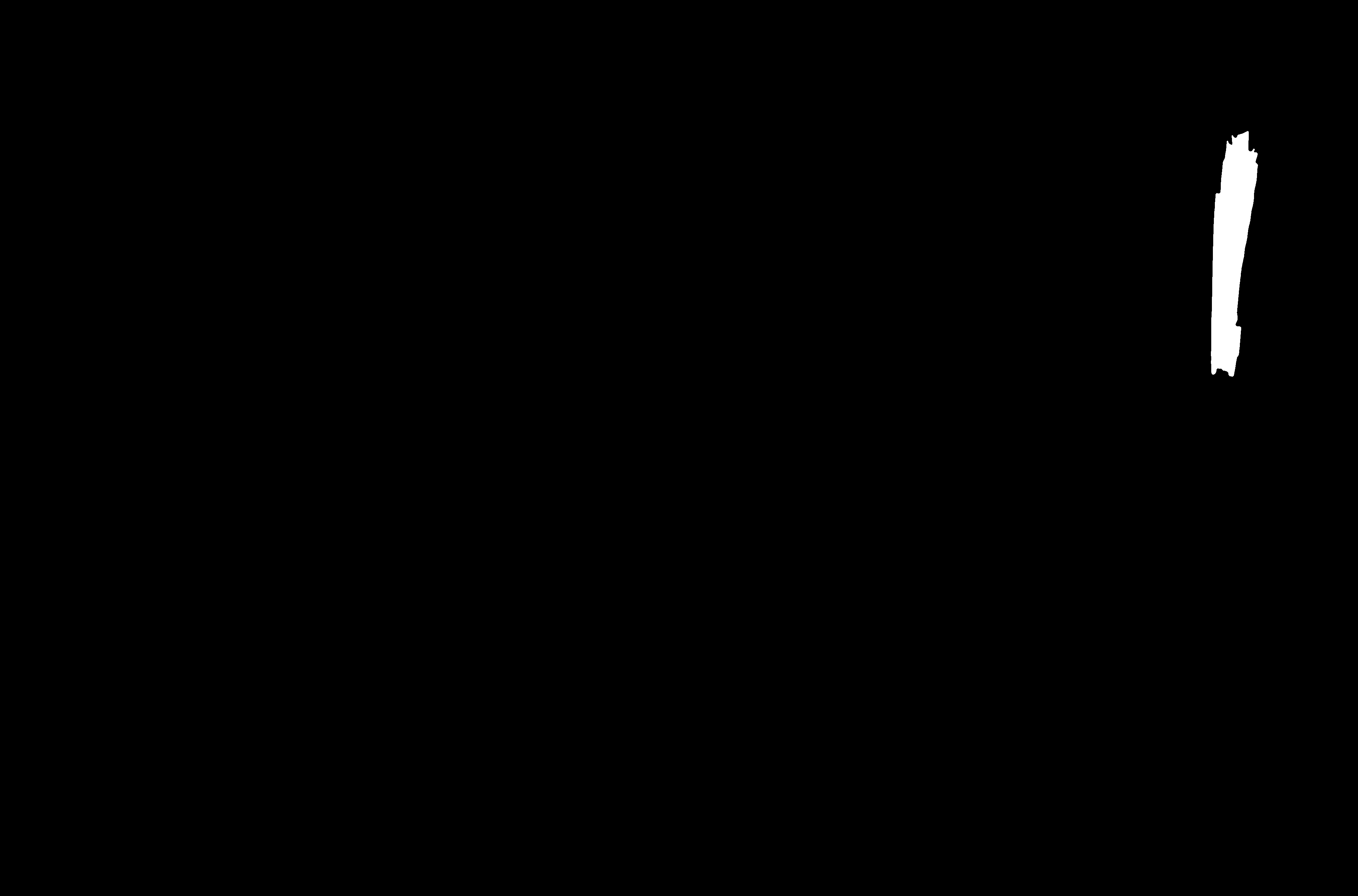}} \\    
  \caption{(a) and (e) Two images from the BIG dataset. (b) and (f) Ground truth semantic segmentation of both the images illustrating the classes dog and potted plant respectively. (c) and (g) Seeds selected by our pipeline. (d) and (h) The final segmentation results obtained by our pipeline using DeepLabV3+ as the base method on the corresponding low-resolution image and Dollar gradient for constructing edge weights. In (c) as there are no seeds selected from bottom portion of the dog, our method fails to reconstruct the bottom portion. As (f) contains a thin object, most of the pixels from the object are discarded by thinning and pruning operators. Thus the number of seeds in (g) are less and the reconstruction in (h) is not satisfactory.}
  \label{fig:Limitations}
\end{figure*}

\paragraph{Dependency on Method used for Low-Resolution Estimates} One of the drawbacks of our pipeline is the dependency on the model used on the corresponding low resolution images for seed selection. This is evident from table \ref{tab:IoU} as the results obtained by refinement on FCN-8 is inferior to the results obtained by refinement on DeepLabV3+. We visualize one such case in the first row of Fig \ref{fig:Limitations}. The first image is an input image from BIG dataset. The second image corresponds to the ground truth segmentations of the class `dog' vs `non-dog' highlighted in white and black respectively. The third column highlights the seeds selected by our pipeline corresponding to `dog'. Observe that the bottom portion of the dog is disconnected from the top portion w.r.t. 4-adjacency in the ground truth. As there are no seeds selected from the second connected component, the random walker fails to construct the bottom part of the dog. Similar comments can be made on small objects. In other words, if the crude segmentation obtained on the low resolution image cannot identify small objects, our pipeline cannot reconstruct such objects. On the other hand, if the smaller objects are captured by the crude segmentation on the low resolution image, our pipeline can reconstruct the objects as all the object instances are preserved by the thinning and pruning operators used in our pipeline. 

\paragraph{Constructing Objects with Low Isoperimetric Quotient}
Recall that the isoperimetric quotient of an object is the ratio of the area of the object to the maximum area of a shape with the same perimeter. Objects with holes, thin objects etc. are typical examples of objects with low isoperimetric quotient. Our method is expected to work well only when the isoperimetric quotient of an object is reasonably high. For example, in the second row of Fig \ref{fig:QualitativeComparison}, our method misclassified the background pixels between the left arm and the left bicep as the background in the local patch has low isoperimetric quotient. The primary reason in this case is that as a result of thinning and pruning, the size of the seeds (of the background) constructed by our pipeline would be very small and insufficient to reconstruct the objects well. The second row in Fig \ref{fig:Limitations} illustrates another such case. The first and second column show the image and the ground truth respectively. Observe that the potted plant is a thin structure. The third column contains the seeds selected by our pipeline. The fourth column show the final segmentation outputs from our pipeline. 

\section{Conclusions and Perspectives}
\label{sec:Conclusions}

In this article, we proposed a pipeline to efficiently extend semantic segmentation algorithms on low-resolution images to high-resolution images. Our approach avoids the bottleneck faced by the majority of the existing methods i.e. the requirement of the ground truth annotations of the high resolution image. Also, our pipeline has very few hyperparameters that are easy to tune. We established that our method is robust to the hyperparameters and it achieves comparable results to the existing state-of-the-art algorithms. Further, we characterize some necessary conditions under which our pipeline is applicable and provide an in-depth analysis of the proposed approach.

The semi-supervised seed propagation algorithm in our pipeline can be replaced with other alternatives. For example, spanning forest-based methods such as watershed cuts \cite{DBLP:journals/pami/CoustyBNC09}, shortest path-based methods such as image foresting transform \cite{falcao2004image} etc are worth exploring. The ideas in the article can be extended to images in other domains by replacing the Dollar gradient with a gradient learnt on the domain-specific images \cite{wibisono2020fined,su2021pixel}. 

Using the seed selection block of our pipeline, a high resolution image can be visualized as a superpixel graph with each superpixel being either a connected component of seeds or an individual non-seed pixel of the high resolution image. One can design neural network architectures on the superpixel graphs to learn the edge weights between adjacent superpixels to obtain semantic segmentation. Further, these ideas can be extended to other computer vision tasks on high resolution images. For example, based on the key insight that boundary pixels are important than the interior pixels, computational complexity of existing algorithms can be improved by non-uniformly downsampling the input images.

\section*{Acknowledgments}

Siddharth Saravanan would like to thank Computer Science and Information Systems, BITS-Pilani Goa. Sravan Danda would like to acknowledge the funding received from BPGC/RIG/2020-21/11-2020/01 (Research Initiation Grant), GOA/ACG/2021-22/Nov/05 (Additional Competitive Grant) both provided by BITS-Pilani K K Birla Goa Campus and thank APPCAIR, and Computer Science and Information Systems, BITS-Pilani Goa.  Aditya Challa would like to thank APPCAIR, and Computer Science and Information Systems, BITS-Pilani Goa.

\bibliographystyle{plain}

\bibliography{references}

\begin{IEEEbiography}[{\includegraphics[width=1in,height=1.25in,clip,keepaspectratio]{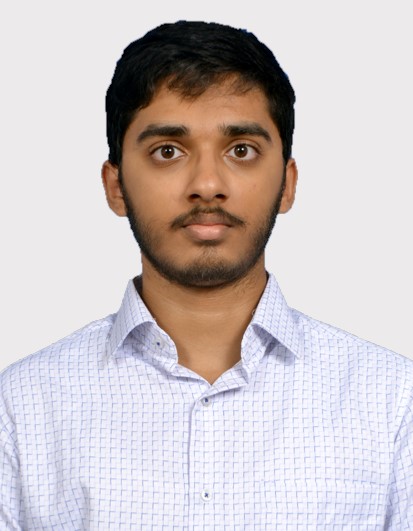}}]{Siddharth Saravanan}
received the B.E. degree in Computer Science with distinction from BITS Pilani, K. K. Birla Goa Campus in 2022. He has interned at CSIR-CEERI, Pilani, and at the Samsung R$\&$D Institute – Bangalore as part of its AR Vision Lab in 2020 and 2022 respectively. He is currently pursuing an M.S. Degree in Computer Science and Engineering from the University of California, San Diego. His current research interests include Computer Vision and Graphics. 
\end{IEEEbiography}

\begin{IEEEbiography}[{\includegraphics[width=1in,height=1.25in,clip,keepaspectratio]{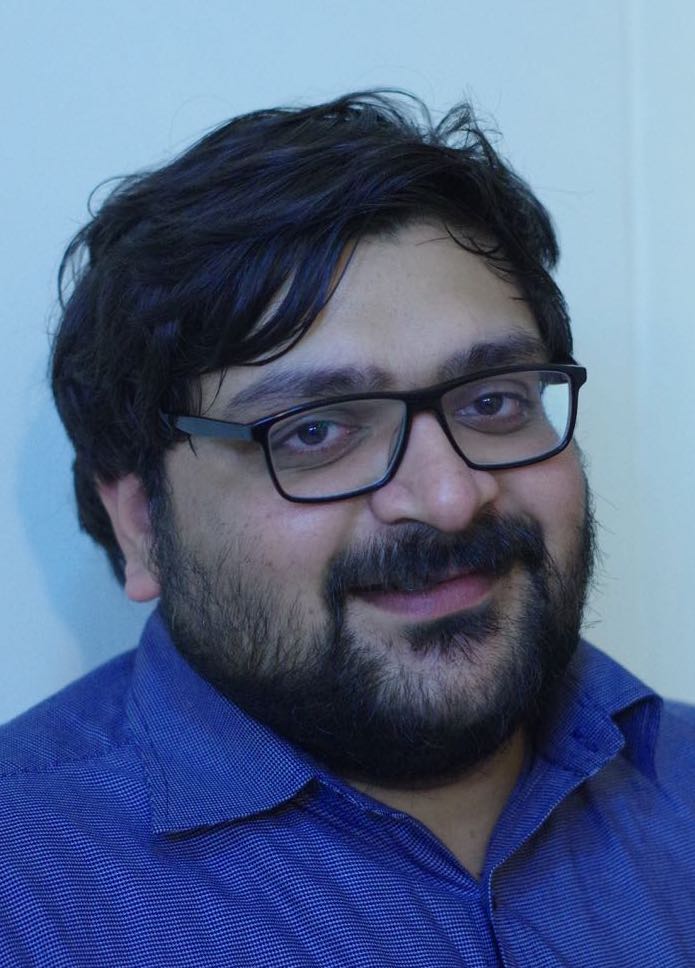}}]{Aditya Challa}
received the B.Math.(Hons.) degree in Mathematics from the Indian Statistical Institute - Bangalore, and Masters in Complex Systems from University of Warwick, UK - in 2010, and 2012, respectively. From 2012 to 2014, he worked as a Business Analyst at Tata Consultancy Services, Bangalore. He completed his PhD in computer science from Systems Science and Informatics Unit, Indian Statistical Institute - Bangalore in 2019. From 2019 to 2021, he worked as a Raman PostDoc Fellow at Indian Institute of Science, Bangalore. He is currently working as an Assistant Professor at Department of Computer Science and Information Systems, BITS Pilani K K Birla Goa Campus.
\end{IEEEbiography}

\begin{IEEEbiography}[{\includegraphics[width=1in,height=1.25in,clip,keepaspectratio]{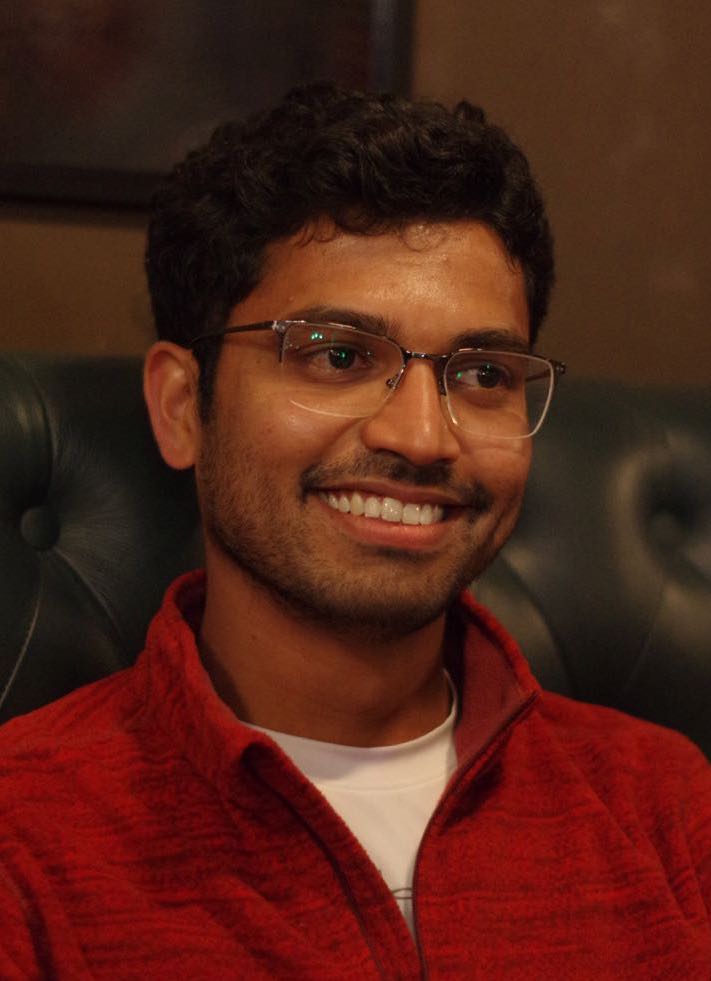}}]{Sravan Danda}
 received the B.Math.(Hons.) degree in Mathematics from the Indian Statistical Institute - Bangalore, and the M.Stat. degree in Mathematical Statistics from the Indian Statistical Institute - Kolkata, in 2009, and 2011, respectively. From 2011 to 2013, he worked as a Business Analyst at Genpact - Retail Analytics, Bangalore. He completed his PhD in computer science from Systems Science and Informatics Unit, Indian Statistical Institute - Bangalore in 2019. He is currently working as an Assistant Professor at Department of Computer Science and Information Systems, BITS Pilani K K Birla Goa Campus. His current research interests are Discrete Mathematical Morphology and Discrete Optimization in Machine Learning.
\end{IEEEbiography}




\end{document}